\pdfoutput=1

\documentclass[11pt]{article}

\usepackage[]{emnlp2021}

\usepackage{times}
\usepackage{latexsym}

\usepackage[T1]{fontenc}

\usepackage[utf8]{inputenc}

\usepackage{microtype}
\usepackage{url}
\usepackage[T1]{fontenc}
\usepackage{amsmath}
\usepackage{amssymb}
\usepackage{tabularx}
\usepackage{mathtools}
\usepackage{booktabs}
\usepackage{longtable}
\usepackage{tabu}
\usepackage{multirow}
\usepackage{amsfonts}
\usepackage{algorithm}
\usepackage{bbm}
\usepackage{subfigure}
\usepackage[noend]{algpseudocode}
\usepackage[normalem]{ulem}
\usepackage{enumitem}
\usepackage{tikz}
\usetikzlibrary{shapes.geometric}
\usetikzlibrary{patterns}
\usepackage{pgfplots}
\usepackage{pgfplotstable}
\pgfplotsset{compat=newest}
\usetikzlibrary{backgrounds}
\usetikzlibrary{matrix,calc}
\def\BState{\State\hskip-\ALG@thistlm}
\usepackage{bm}
\usepackage{xcolor}
\DeclareMathOperator*{\argmax}{arg\,max}

\newcommand{\xv}{\ensuremath{\bm{x}}}
\newcommand{\yv}{\ensuremath{\bm{y}}}

\usepackage[T2A,LGR,T1]{fontenc}
\usepackage[utf8]{inputenc}
\usepackage[russian,greek,spanish,es-nodecimaldot, english]{babel}
\usepackage[normalem]{ulem}

\title{When is \textit{Wall} a \textit{Pared} and when a \textit{Muro}?:\\Extracting Rules Governing Lexical Selection }

 \author{Aditi Chaudhary$^\dagger$, Kayo Yin$^\dagger$, Antonios Anastasopoulos$^\ddagger$, Graham Neubig$^\dagger$ \\
   $^\dagger$Carnegie Mellon University, $^\ddagger$George Mason University \\
    \texttt{\{aschaudh,kayoy,gneubig\}@cs.cmu.edu} \hspace{.5cm} \texttt{antonis@gmu.edu}}

\begin{document}
\maketitle
\begin{abstract}
Learning fine-grained distinctions between vocabulary items is a key challenge in learning a new language. For example, the noun ``wall'' has different lexical manifestations in Spanish --  ``pared'' refers to an indoor wall while ``muro'' refers to an outside wall.
However, this variety of lexical distinction may not be obvious to non-native learners unless the distinction is explained in such a way.
In this work, we present a method for \emph{automatically} identifying fine-grained lexical distinctions, and extracting concise descriptions  explaining these distinctions in a human- and machine-readable format. We confirm the quality of these extracted descriptions in a language learning setup for two languages, Spanish and Greek,  
where we use them to teach non-native speakers when to translate a given ambiguous word into its different possible translations. Code and data are publicly released here.\footnote{\url{https://github.com/Aditi138/LexSelection}}
\end{abstract}

\section{Introduction}

\begin{figure}[t]
\centering 
\includegraphics[width=.9\columnwidth]{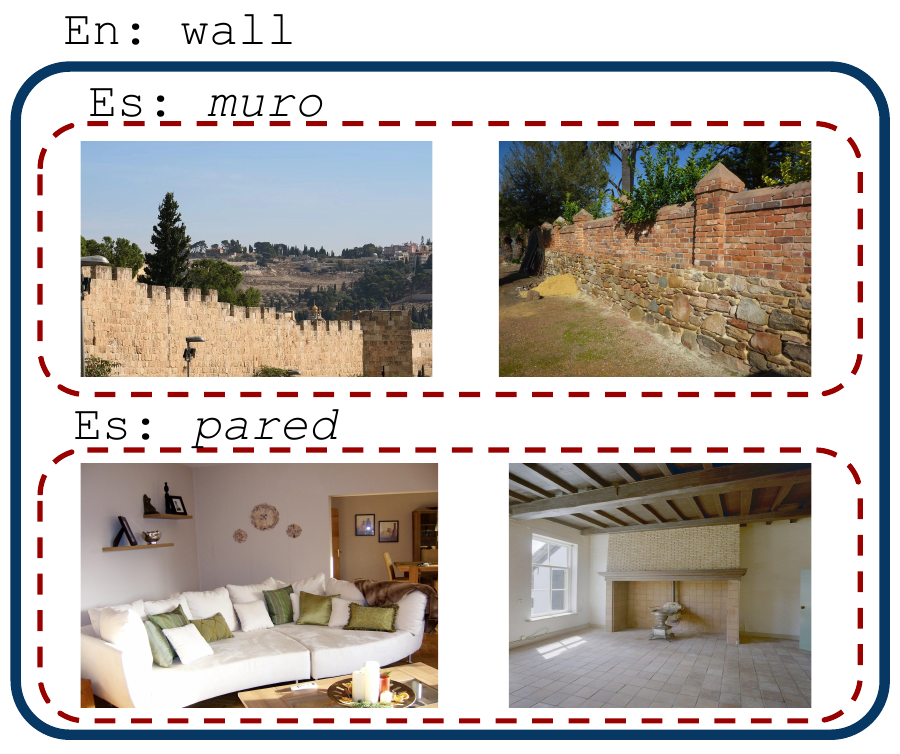}

\caption{Semantic subdivision for the concept `wall' results in different lexical manifestations in Spanish: `muro' for \emph{outside wall} and `pared' for \emph{inside wall} whereas in English both are referred as `wall'.}
\label{fig:overview}
\vspace{-1em}
\end{figure}

With increasing globalization there is a widespread prevalence and need for good materials and tools to help people learn languages. Curating such content manually requires a large time and cost investment which poses a challenge particularly for languages where protection and revival efforts are ongoing \cite{moline2020indigenous}.
One of the most important and challenging processes in learning a new language (L2) is vocabulary acquisition \cite{ellis1996sequencing,moore1996foreign}, which is generally made easy by associating L2 words with words from the first language (L1) \cite{hulstijn1996incidental,watanabe1997input}.
In many cases, L1 words or word senses can be unambiguously associated with L2 words.
For example ``linguistics'' and ``ling\"{u}\'{i}stica'' essentially form a one-to-one mapping between English and Spanish.
However, different languages carve up the semantic space of the world in different ways leading to \emph{semantic subdivisions}, distinctions made in one language not made in another.
For example, ``wall'' in English is manifested differently in Spanish as ``pared'' or ``muro'', as shown in Figure \ref{fig:overview}, and for an L1 English speaker it may not be immediately obvious when one should be used over the other. A skilled teacher or comprehensive language learning resource may be able to provide explanations that resolve this ambiguity. For example, \citet{robertson-2020-show} present word definitions in-context for Finnish learners, while CAVOCA \cite{groot2000computer} takes a learner through various stages of the word acquisition process including word usage,  syntax.

In this work, we propose a method to \emph{automatically} discover rules regarding fine-grained lexical distinctions and present L2 learners with concise descriptions derived from them in an interactive framework. 
Research in L2 vocabulary acquisition \cite{groot2000computer} 
has shown that it is effective to combine strategies using explicit definitions and examples in context. 
However, our work contrasts to most prior work in this field such as CAVOCA \cite{groot2000computer} or Duolingo,%
\footnote{\url{https://www.duolingo.com/}}
which use learning content manually created by subject matter experts. This necessity for curation makes it difficult to comprehensively scale this approach to many languages. 

Specifically, our framework consists of two steps: i) use a parallel corpus to identify words in L1 which have different lexical manifestations owing to a semantic subdivision in  L2, and ii) create  human- and machine-readable concise descriptions  that allow for easier interpretation of each lexical distinction. First, we extract source (L1) and target (L2) parallel sentences for each shortlisted L1 word. We then extract lexical and semantic features, as well as a label encoding the lexical choice in the target language from these parallel sentences for each L1 word. Finally, we train a prediction model that distinguishes between the lexical choices, and extract human-understandable descriptions from this model.  These descriptions could either be used as-is, or could be used as a starting point for further curation by educators. 

To confirm the quality of the extracted descriptions, we conduct a study where we use them to teach English native speakers lexical distinctions arising from semantic subdivisions in Spanish and Greek. We make our study interactive by presenting the learning content in the form of \emph{cloze} tests \cite{taylor1953cloze} where the English word to be taught is presented to the learner in context along with extracted concise description. The learner is then required to select the most appropriate lexical choice from the given set. 
The main methodological contributions therefore are \emph{automated} methods to:
\begin{itemize}[noitemsep,leftmargin=*]
    \item Identify fine-grained lexical distinctions arising due to semantic subdivisions. To evaluate this and future work, we also create a lexical selection dataset for two language pairs, English-Spanish and English-Greek.
    \item Extract rules to help humans understand the usage of lexical distinctions in context. 
    Studies with 7 Spanish  and 9 Greek learners show that they learn faster when given access to our extracted descriptions; for example they achieve an (avg.) accuracy of 81\!\% within roughly 20 questions, as opposed to more than 40 questions required otherwise. 
\end{itemize}


\section{Problem Formulation}
For the purpose of this paper, we define the task of \emph{lexical selection} as choosing contextually correct translations from a set of target translations for an ambiguous word in the source language  \cite{lefever-hoste-2010-semeval}.
We first define some variables: $\xv=x_1,x_2,\ldots,x_{|\xv|}$ denotes a sentence in the source language (L1),  $\yv=y_1,y_2,\ldots,y_{|\yv|}$ is its translation in the target language (L2) and $V_x$ and $V_y$ are the source and target vocabulary respectively. 
Given a source sentence $\xv$  containing an ambiguous word $x_i$, then $\text{trans}(x_i) \subseteq V_y$ denotes the set of its ``possible'' target translations  i.e. words in the target language to which the ambiguous word $x_i$ might be translated (concrete methods to define this set are explained later). 
The task of lexical selection involves choosing the most appropriate translation $y_i \in \text{trans}(x_i)$, and can be performed either by machines or humans.\footnote{The notation here refers to single-word translations which are the focus of this work.} In this work, we particularly focus on \emph{machine-learned methods to help humans learn lexical selection}, extracting lexical selection models that are not only usable by machines, but also interpretable by humans in order to aid the process of learning a new language. We thus plan to extract the rule set $\mathcal{R}_{v_x}$ which governs this lexical selection process in a human- and machine-readable format.

\section{Identifying Semantic Subdivisions}
\label{sec:ambig}
In this section, we describe in detail the procedure for identifying L1 words that have  different lexical manifestations in L2 owing to semantic subdivisions. For the purpose of this work, we refer to these different lexical manifestations in L2 as lexical choices and the corresponding L1 words as focus words.
Our work is ``loosely inspired'' by  ContraWSD \cite{rios-etal-2018-word} and SemEval-2013 \cite{lefever-hoste-2013-semeval} which construct a dataset for  cross-lingual word sense disambiguation, using a semi-automatic approach combining frequency-based heuristics  with  human supervision. 
These datasets are restricted to  a subset of manually selected nouns (20  for SemEval-2013 and 70-80 for ContraWSD). 
In contrast, our approach is fully automated going beyond using just frequency-based filters. Furthermore, we do not restrict to any one word class leading to  words being identified across different word classes (nouns, verbs, adjectives, adverbs) for both Spanish and Greek.\footnote{More details in Section \S\ref{ref:automated}.}

We start with a parallel corpus  $D\!=\!\{(\xv_1, \yv_1), \cdots, (\xv_{|D|},\yv_{|D|})\}$  where $(\xv_m,\yv_m)$ denote the source and target sentence pair. Next, we extract word alignments automatically using a word aligner that finds sets of pairs of source and target words
$A_m=\{ \langle x_i, y_j\rangle: x_i \in \xv_m, y_j \in \yv_m \}$, where for each word pair $\langle x_i, y_j\rangle$, $x_i$ and $y_j$ are semantically similar to each other within this context. 

To focus on translations of the underlying content, as opposed to morphological variations, we then lemmatize all words in both the source and target sentence pairs. Thus, $V_x$ and $V_y$ refer to the lemmatized vocabulary of the source and target language.
Going forward, all words refer to  their respective lemmatized forms.
We  perform automatic part-of-speech (POS) tagging, dependency parsing  and word sense disambiguation (WSD) on the source side data, resulting in a POS tag and word sense associated with each source word, $\text{tag}(x_i) \in T_x$ and $\text{sense}(x_i) \in S_x$ where $T_x$ is the set of POS tags and $S_x$ is the word sense vocabulary in the source language.

In order to identify the focus words, we extract a list of lemmatized L1 word types $v_x$ filtered by their part-of-speech (POS) tags $t_x$ giving us tuples of the form $\langle v_x, t_x \rangle$. This ensures that we don't conflate meanings across  POS tags, because in many languages the semantics of a word can vary widely across its different POS tags.\footnote{``Brown'' as a verb (as in ``brown the meat'') is treated differently from the adjective sense (as in ``brown hair'').}
We refer to the extracted tuples $\langle v_x, t_x \rangle$ as focus words for simplicity. We then extract the focus words with their respective lexical choices as follows: 

\paragraph{1.~Extract translations}: For each aligned word pair $\langle x_i, y_j \rangle$ 
    compute the number of times $c(v_x,t_x,v_y)$ the lemmatized source word type  $(v_x\!=\!\text{lemma}(x_i))$ along with its POS tag $(t_x\!=\!\text{tag}(x_i))$ is aligned to the lemmatized target word type $(v_y\!=\!\text{lemma}(y_j))$ across the whole corpus. Also, store the number of times the word sense of $x_i$ ($s_x\!=\!\text{sense}(x_i)$) appears with the source word type, source POS tag and the translation word type in $g(v_x,t_x,s_x,v_y)$.
    
    \paragraph{2.~Filter on frequency}: Extract tuples of source types and POS tags  $\langle v_x, t_x\rangle$ that have been aligned to at least two target words at least 50 times ($\{v_y:|c(v_x,t_x,v_y) \ge 50\}| \ge 2$), to account for alignment errors. To avoid ambiguity on the target side, translations aligned to words other than the word $v_x$ in question (at least 3 times) are excluded.
    
    \paragraph{3.~Filter on entropy}: Remove source tuples that have an entropy $H(v_x, t_x)$ less than a pre-selected threshold. The entropy is computed using the  conditional probability of a target translation given the source type and POS tag:
    \[ p:= p(v_y|v_x, t_x) =\frac{c(v_x,t_x,v_y)}{c(v_x,t_x)} \]
\[ H(v_x, t_x) = \sum_{v_y \in \text{trans}(v_x, t_x)}  -p\log_e p \] where  $\text{trans}(v_x, t_x)$ is the set of target translations for the source tuple $\langle v_x, t_x \rangle$ and $p(v_y|v_x, t_x)$ is the conditional probability of the target translation for this source type $v_x$ and its POS tag $t_x$. High entropy suggests that a word is ambiguous, with fine-grained distinctions that likely require context to be resolved, and thus is a word we should focus on.

    \paragraph{4.~Filter on word sense}: Remove source tuples whose target translations have distinct source-word senses. For some words, the differences between target translations can be straightforwardly explained by the different source word senses. For example, \textit{banco} in Spanish refers to the financial institution, given by the WordNet \cite{miller1995wordnet} sense `bank.n.02' while \textit{orilla} refers to the edge of a river, outright matched to `bank.n.01'. For such words, the word sense definitions would be an easy-to-provide rule for learners, but we want to go beyond that. We are interested in finding those  words where the word sense information alone is insufficient to distinguish between the lexical choices and are hence likely to be hard for human learners.
    For a source tuple, use the highest occurring word sense for a given target translation $v_y$ computed as:
    \[ Q(v_y)\!=\!\argmax_{s_x \in S_x} g(v_x,t_x,s_x,v_y)\]
    
    Finally, retain the source tuples whose target translations all have the same sense, giving us $L$ lexical choices $\text{trans}(v_x,t_x) = \{ v_{y_0}, \ldots, v_{y_{|L|}}\}$ for a source tuple $(v_x, t_x)$.
    


\section{Lexical Selection Model}

After identifying a set of focus words in the source language, we train a lexical selection model parameterized by  $\theta_{\langle v_x, t_x \rangle}$ for each focus word $\langle v_x, t_x \rangle$. 
We extract the parallel sentences from $D$ that include the focus word and its corresponding lexical choices, denoting them with $D_{\langle v_x, t_x\rangle}$. 
The model takes as input the  source sentences  $\xv_{\langle v_x, t_x \rangle} \in D_{\langle v_x, t_x \rangle}$   and predicts the contextually correct target translation $v_y$ from a set of possible translations $\text{trans}(v_x, t_x) = v_{y_1}, v_{y_2}, \cdots, v_{y_k}$

Since we aim to induce concise, human-understandable explanations of semantic distinctions that can be presented to learners to help them better understand the lexical selection process, we train a prediction model which allows us to easily extract such descriptions for each lexical choice $v_y \in \text{trans}(v_x, t_x)$.  In this paper, we use human-readable descriptions of the features learned by a linear model, where these features are defined over a set of lexical and semantic features extracted from the source sentences  in $D_{\langle v_x, t_x \rangle}$.
For designing  features, we take inspiration from prior work which uses extracted contextual information to improve cross-lingual sense disambiguation in machine translation systems~\cite{garcia2001refined,carpuat2007improving,carpuat2007phrase}. 


\subsection{Model Features}
For training a lexical selection model $\theta_{\langle v_x, t_x \rangle}$ for the focus word $\langle v_x, t_x \rangle$, we construct training data from the source-target sentence pairs $D_{\langle v_x, t_x \rangle}$.  We focus on features extracted only from the current source sentence, although the framework can be easily extended to include features from the target sentence as well.
We represent each source sentence $\xv_{\langle v_x, t_x \rangle}\!\in\!D_{\langle v_x, t_x \rangle}$  with a set of features extracted from the \textit{neighborhood} of the focus word context relevant to the lexical selection process. This neighborhood includes (1) words from the source sentence that occur within a fixed window of the given ambiguous word, and (2) the head and dependents of the  focus word as given by the dependency parse of the sentence. For each word in this relevant context, we extract the following lexical features:
\begin{itemize}[noitemsep,nolistsep,leftmargin=*]
    \item \textbf{Lemma} Lemma of the token.
    \item \textbf{WSD} Word sense of the token as extracted from a state-of-the-art word sense disambiguation (WSD) model.
    \item \textbf{Bigram} Bigrams constructed from lemmas of the words present within a fixed window around the focus word. We exclude punctuation and stop words within the window.\footnote{Stop words as provided by NLTK\cite{nltk}}
\end{itemize}

\subsection{Model Training}
To enable extraction of human-understandable descriptions, we  use a model that is conducive to interpretation: the linear SVM \cite[LinearSVM;][]{cortes1995support}, which gives us feature weights $\theta_{\langle v_x, t_x \rangle}$ that can be easily interpreted as the importance of each feature in making the decision. %
Since there can be $n$-ary lexical choices for a given focus word, we train using the one-vs-rest (OvR) method which trains one model per each lexical choice $v_{y_k}$, where data from ${v_{y_k}}$ are treated as positive examples and data from all other choices as negative, allowing us to extract feature weights for each decision.

\begin{figure*}%
\centering
\includegraphics[width=0.9\textwidth]{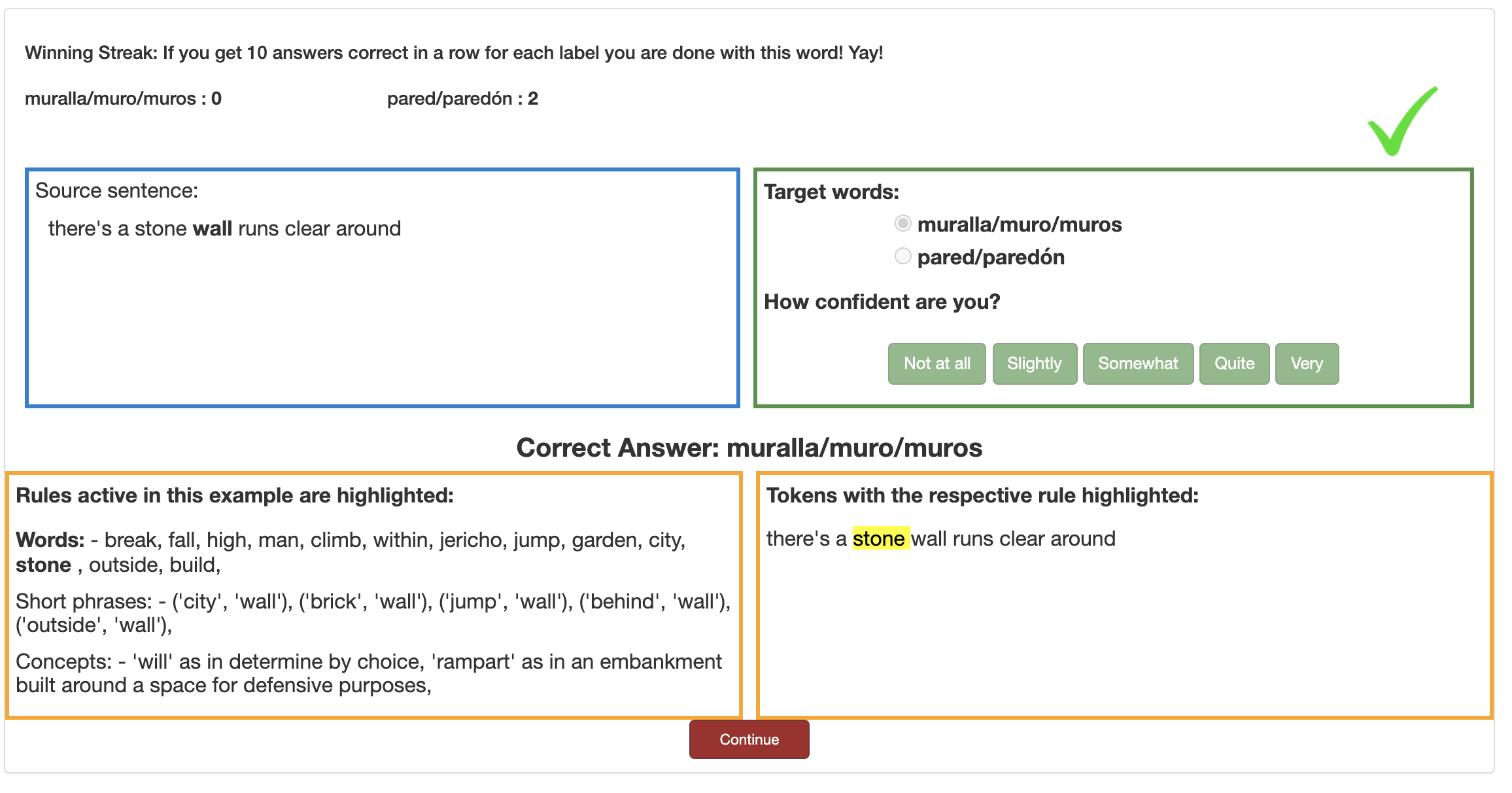}
  \caption{ Learning Interface. Rules for the correct answer are displayed  to the learner after each question. Individual rules that apply to the given example are highlighted for the convenience of the learner. ``wall'' here refers to an outside wall and the adjective \emph{stone} serves as a hint in arriving at the correct answer. }%
\label{fig:taskb}%
\vspace{-1em}
\end{figure*}

\subsection{Rule Extraction} 
As mentioned above, we use human-readable descriptions of the features learned by a linear model to be presented to the human learners. More broadly, we refer to these descriptions as ``rules'', however these rules could take other forms as well, and we hope that future work by us or others could find other creative ways to induce or define these rules.

For each focus word $\langle v_x,t_x \rangle$, we extract the rule set $\mathcal{R}_{\langle v_x,t_x,v_{y_k} \rangle}$, which is the set of rules for selecting a given lexical choice $v_{y_k}$ from the set of possible choices $\text{trans}(v_x, t_x)$. For this, we extract salient features from the trained model $\theta_{\langle v_x, t_x \rangle}$ for each lexical choice. As mentioned above, using the OvR classification method we get one model per choice  $v_{y_k}$, from which we can then extract the top-$N$ features  having the highest weight coefficients for each choice. In order to present this rules in a human-readable form, we create concise rule templates as shown in Appendix \ref{app:templates}. 

\section{Automated Validation}
\label{ref:automated}
Since our main research goal is to aid human learners in their learning, we focus on two approaches of evaluation:
(a) automated validation, a preliminary evaluation where we validate to what extent our interpretable model can perform cross-lingual lexical selection, and (b) human evaluation (\S \ref{sec:humanexpt}) which answers our main question of whether it can teach human learners the usage of L2 words.


For the automated evaluation in particular, we verify several things.
First, we check whether our interpretable lexical selection model is able to learn cross-lingual lexical selection at all by measuring its performance compared to selecting the most frequently occurring translation in the corpus for a given focus word (``Frequency'').
We also compare with another alternative interpretable model, decision trees (DTree) trained using the same features as LinearSVM, to validate the choice of SVMs as an interpretable model over other alternatives. 
Further, we check how our interpretable linear SVM model compares with a ``performance skyline''; a less interpretable BERT-based neural model \cite{devlin2018bert} that extracts representations of the source sentence from BERT and trains a classifier to predict the correct lexical choice.

\subsection{Setup}

\paragraph{Data:} We experiment with two L2 languages: Spanish and Greek. These languages were chosen due to (1) availability of parallel corpora with which to train models, and (2) availability of linguists and annotators to verify and analyze the data used in our experimental setting.
For Spanish we use 10 million English-Spanish parallel sentences from OpenSubtitles~\cite{lison-tiedemann-2016-opensubtitles2016}, Tatoeba, TED~\cite{TIEDEMANN12.463}, and Europarl \cite{koehn2005epc}.
 \footnote{We use only 1 million sentences from Europarl because we found sentences from Europarl to contain fewer semantic subdivisions owing to the very specific domain of the dataset.} 
For Greek, we use 31 million English-Greek parallel sentences extracted from OpenSubtitles.
For word alignment we use the AWESOME aligner \cite{dou2021word}, for lemmatization we use spaCy \cite{spacy}, for POS tagging and dependency parsing we use Stanza \cite{qi-etal-2020-stanza}, and for English WSD we use EWISER \cite{bevilacqua-navigli-2020-breaking}.%
\footnote{POS tagging, dependency parsing and WSD is required \emph{only} for the source language, here English.}

Using our automatic pipeline (\S \ref{sec:ambig}), we identify 157 English words which have  fine-grained  distinctions in Spanish and 707 English words for Greek. Among these,
for Spanish there are 127 nouns, 15 verbs, 10  adjectives, 5 adverbs and, for Greek there are 452 nouns, 123 verbs, 126 adjectives and 6 adverbs. Along with nouns which do account for much of the data, we do find significant number of verbs and adjectives also exhibiting $\geq2$ lexical choices.\footnote{More details in Appendix \ref{app:auto}} 
A manual inspection by a Greek-English bilingual speaker revealed that most automatically created lexical choices were correct. In just a couple of cases, lemmatizer errors lead to two choices corresponding to the same actual lemma (which were manually corrected for the user studies).

\paragraph{Model:}
We train a linear SVM lexical selection model with \texttt{sklearn} \cite{pedregosa2011scikit} for each  L1 focus word  and divide the extracted parallel sentences into a train/test split with a 80-20 ratio per lexical choice. 
We perform 5-fold cross-validation to select the best model hyperparameters (detailed in Appendix \ref{app:setup}) from which we then extract the top-20 features for each lexical choice to form our rule set. 
Details on the  setup of DTree and BERT are also in Appendix \ref{app:setup}.

\subsection{Results}
Table \ref{tab:model_data} shows the test accuracy averaged across all focus words for both Spanish and Greek.
Regarding our underlying questions, we first find that LinearSVM significantly outperforms both Frequency and DTree by a significant margin, indicating that it is both learning to perform lexical selection to a significant degree, and outperforming other reasonable alternatives for interpretable models.%
\footnote{Individual scores per focus word listed in Appendix \ref{app:setup}}
This gives us confidence to proceed to use it in our following human learning experiments.
Interestingly, our interpretable LinearSVM model is within 97\% relative accuracy of the skyline BERT model (just 2.09 percentage points behind). 
The fact that the more complicated but less inherently interpretable BERT model is better overall paves the way for future work in applying model interpretation techniques \cite[\textit{inter alia}]{abnar-zuidema-2020-quantifying} to extract human-interpretable rules for lexical selection, although this is beyond the scope of the current paper.\footnote{Overall accuracy is low, with even BERT getting 70\%,  possibly due to lack of sufficient source-side context. OpenSubtitles comprises of movie dialogues where the sufficient context could span more than a single sentence.}
We find that lexical selection accuracy varies by part of speech; all models perform poorly on adverbs with (avg.) gain of only +0.97 points over the baseline (c.f. with gains of +8.04 for nouns, +5.16 for verbs, +6.24 for adjectives).

 \begin{table}[t]
\small
    \centering
    \resizebox{\columnwidth}{!}{
    \begin{tabular}{c|l|c|c|c|c|c}
    \multirow{2}{*}{\textbf{Lang.}} & \multirow{2}{*}{\textbf{Model}} & \multicolumn{5}{c}{\textbf{Test Accuracy}} \\
   &  & All & nouns & verbs & adj.  & adv. \\ 
   \midrule
    \multirow{4}{*}{Spanish} 
     & Frequency (Baseline) & 59.43 & 59.36 & 60.17 & 60.67  & 53.03 \\   
     & DTree & 62.40 & 62.45 & 61.57 & 65.22  &  54.82\\   
     & LinearSVM & \textbf{66.87} & \textbf{67.41} & \textbf{65.34} & \textbf{66.91} & \textbf{56.29}  \\    
     & BERT & \underline{70.72} & \underline{71.75} & \underline{69.04} & \underline{67.31}& \underline{54.07} \\    
     \midrule
     \multirow{4}{*}{Greek} 
     & Baseline & 58.56 &  59.48&  53.04& 60.48 & 61.82 \\   
     & DTree & 63.79 & 64.49 &  59.74 & 65.39  & 61.13  \\    
     & LinearSVM & \textbf{66.46} & \textbf{67.09} & \textbf{63.30}  & \textbf{67.51} & \textbf{64.98} \\    
     & BERT & \underline{71.74} & \underline{70.91} & \underline{78.14} & \underline{68.86} &  \underline{62.76}\\    
    \bottomrule
    \end{tabular}
    }
    \caption{The interpretable LinearSVM lexical selection model is almost on par with the BERT \underline{\smash{skyline}}.}
    \label{tab:model_data}
    \vspace{-1em}
\end{table}

\section{Evaluation with Human Learners}
\label{sec:humanexpt}
We move to our main evaluation where we examine \emph{how effective our extracted rules are in aiding human learners} in understanding the distinctions in L2 words. 

\subsection{Evaluation Methodology}
We take inspiration from  existing research on second language acquisition (SLA)  to design our evaluation method.
For instance, \citet{groot2000computer}  highlights the different learning strategies which are based on generally accepted language acquisition theories \cite{nation2005teaching,richards1999exploring}, which suggest that a learner  is required to go through different levels of language processing for effectively learning vocabulary. 
In particular, \citet{groot2000computer} empirically show  that some of these levels can be accelerated with appropriate design of the language tasks  by combining learning strategies which use both examples in context and definitions for effective learning. 
Our cloze-style tasks are essentially examples in context showing the word usage in a given context and the extracted rules are a proxy for human-provided definitions. 

Specifically, we set up an interactive exercise where a human learner is presented with the English focus word in context, along with a set of possible L2 (Spanish or Greek) lexical choices. The learner is then required to select one of the possible lexical choices, based on which they think correctly translates the focus word in the given source context. They must also mark  how confident they are in their answer (``Not at all'', ``Slightly'',  ``Somewhat'', ``Quite'' or ``Very''). After they select the answer to each question, they are told the correct answer immediately. For each focus word, we ask the learner to answer up to $N$ multiple-choice questions in sequence, which contain roughly equal number of questions for each lexical choice.

In order to evaluate how effective the extracted rules are in aiding the  learning process, we perform this study in two setups, a baseline one without rules, and one using our proposed system with rules.

\paragraph{Baseline Setup:}
In this setup, the human learner does not have access to any rules and immediately starts answering questions. If the learners do not know the target language, they are likely to start out with approximately chance accuracy (e.g.~50\% if there are two choices), but as they are given feedback they may be able to grasp the patterns under which one particular translation or another is used, and gradually rise above chance accuracy.

\paragraph{Proposed Setup:}
In the proposed setup, before starting the task, the learner is shown  brief rules regarding when you would use each possible lexical choice $v_{y_k}\!\in\!\text{trans}(v_x, t_x)$, constructed from the rule set $R_{\langle v_x, t_x, v_{y_k} \rangle}$.
They take as much time as they want to review these rules, and then move to answering questions.
The interface for answering questions is the same as the baseline, but below the task screen they can review the rules of different translation choices (figures in Appendix \ref{app:interface}).
On selecting a choice, the learner is shown the correct answer accompanied with its corresponding human-readable rules of \emph{only} the correct answer. Further, we highlight those individual rules that helped decide the correct answer (Figure \ref{fig:taskb}) for the convenience of the learner. 
By highlighting it in the two bottom panes, we hope to draw the learner's attention to these hints and thus strengthen the understanding of the underlying concept.

In this setting, the annotator may achieve non-chance accuracy even at the very beginning of answering questions, as they have been given an explanation regarding the underlying rules that they can leverage in answering questions.
The accuracy will likely further increase as they practice and become familiar with actual examples and how the extracted features apply to them. 

\subsection{Experimental Details}
We select native English speakers, 7 for the Spanish study and 9 for the Greek study.\footnote{We allow participants who know other languages but none that are familiar with the L2 or its related languages.} Each annotator is presented with the same set of English  words or tasks.
For each study, half of the words will be annotated using the baseline setup and remaining half with the proposed setup. 
To ensure an unbiased setup, we randomize whether each focus word uses rules or not, while ensuring that  at least half the annotators see the proposed setup and the other half perform the same task in the baseline setup for each word.
We  further shuffle the order in which the words are presented. For each English word, we select up to 40 examples each for the respective lexical choices. However, as an incentive, we end a task early if the annotator correctly answers 10 questions straight in a row for each  lexical choice. 
We explain below the selection procedure for the English words used in the  experiments.

\paragraph{Word Selection:} In an ideal situation, we would like to conduct 
these experiments for all identified English focus words, but this would involve annotating thousands of sentences, requiring a large time commitment from the annotators. Instead, we shortlist a handful of words  using the following automated procedure: First, for a given L2 study, we sort  all focus words by the number of  available data points ($D_{\langle v_x, t_x \rangle}$).
Next, from the trained lexical selection model $\theta_{\langle v_x, t_x \rangle}$ we compute an F1-score for each lexical choice and  filter focus words where the model gets an F1 $>0.5$ for each lexical choice. Finally, we select upto 10 focus words with the most data points that fit the above condition. For each word $(\langle v_x, t_x \rangle)$, we then select 40 \textit{representative} examples for each lexical choice (see paragraph below). Details on the shortlisted words  can be found in Appendix \ref{app:lexical}.

\paragraph{Representative Example Selection:} To facilitate  an effective learning process, we present examples to the learner that have  \emph{sufficient source-side context} required for correctly identifying the target-side lexical choice.
This is important because there are examples in the corpus where the sufficient context requires context spanning over multiple sentences. 
To make our learning content both concise and effective,  we focus only on context self-contained in a single sentence. 
Further to efficiently conduct a high-quality study, we enlist help from native  speakers of the L2 language to filter the required sentences. We note, though, that the relevant sentences could also be potentially filtered automatically (left for future work).

To get such meaningful examples, we present bilingual English-Spanish and English-Greek speakers with the English sentence containing the focus word and the set of possible lexical choices in Spanish and Greek respectively. 
They then select the word which best suits the given context and mark their confidence in the selection.
The interface for the example selection is the same as Figure \ref{fig:taskb} (but without rules). We collect these annotations from multiple native speakers and only keep those sentences on which all native speakers agree (see Appendix \ref{app:lexical} for details).

\subsection{Results and Discussion}

To confirm whether the extracted rules are effective to the learning process, we examine the following questions:

\paragraph{Do the extracted rules result in increased learner accuracy?}
We compute the learner accuracy across all learners for each L2 study. If a learner attains higher accuracy with fewer attempted examples for the experiment with rules than without, then the extracted rules could be considered effective in the learning process. 
However, we cannot directly use the learner accuracy as-is because of the possibility of 
other sources of variability such as (a) underlying learner ability, as some learners may be more proficient than others, (b) underlying task difficulty, as some words may be harder to disambiguate than others, or (c) word ordering, as learners may become proficient as they do more tasks. Therefore, we use a mixed effects model \cite{mclean1991unified}, which models \emph{random effects} and \emph{fixed effects} to account for such random variability. Random effects are  variables responsible for random variation such as task-identity, task-order and the learner, while fixed effects such as the presence of rules are the variables of interest for determining the response variable i.e. learner accuracy. A linear mixed-effect model (LME) is defined as:
$\mathbf{y} = \mathbf{X}\mathbf{\beta} + \mathbf{Z}\mathbf{u} + \mathbf{\epsilon}$ where $\mathbf{y}$ is the learner accuracy, $\mathbf{\beta}$ and $\mathbf{u}$ are the fixed-effect and random-effect regression coefficients, $\mathbf{X}$ and $\mathbf{Z}$ are the respective design matrices and $\epsilon$ the noise.

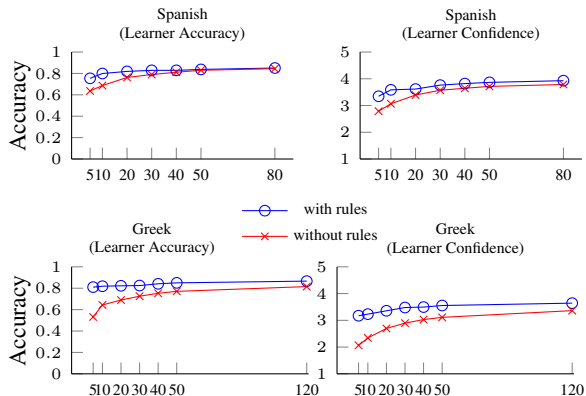
\begin{figure}
\pgfplotstableread[row sep=\\,col sep=&]{
Number & norule & rule\\
5	& 0.63592 & 	0.75412 \\ 
10	& 0.68669 &	0.79922 \\
20	& 0.76223 &	0.81879 \\
30	& 0.78967 &	0.82899 \\
40	& 0.81113 &	0.82857 \\
50	& 0.829532 &	0.837361 \\
80	& 0.844433	 & 0.85051 \\
}\spadata
\pgfplotstableread[row sep=\\,col sep=&]{
Number&	norule &	rule \\
5	& 2.7872 &	3.3454 \\
10	& 3.0661 &	3.5867 \\
20	& 3.3925 &	3.6154 \\
30	& 3.5709 &	3.7631 \\
40	& 3.6454 &	3.8205 \\
50	& 3.7151 &	3.8657 \\
80	& 3.786 & 	3.9293 \\
}\spaconfdata
\pgfplotstableread[row sep=\\,col sep=&]{
Number & norule & rule\\
5	& 0.53175 & 0.81046 \\ 
10	& 0.64342 &	0.81879\\
20	& 0.68949 &	0.82272  \\
30	& 0.72546 & 0.82491 \\
40	& 0.75283 &	0.84137 \\
50	& 0.7707 &	0.85038 \\
    120	& 0.81533	 & 0.8664 \\
}\greekdata
\pgfplotstableread[row sep=\\,col sep=&]{
Number&	norule &	rule \\
5	& 2.0646 &	3.172 \\
10	& 2.3419 &	3.2313 \\
20	& 2.6964 &	3.3565 \\
30	& 2.8947 &	3.4782 \\
40	& 3.0282 &	3.4961\\
50	& 3.1114 &	3.5538 \\
120	&3.3622 & 	3.6424 \\
}\greekconfdata

\begin{tikzpicture}[trim left=-0.85cm,trim right=0cm]
    \begin{axis}[
            every axis plot post/.style={/pgf/number format/fixed},
            width=4.5cm,
            height=3cm,
            ymajorgrids=false,
            yminorgrids=false,
            legend style={draw=none},
            xtick={5,10,20,30,40,50,80},
            xticklabels={5,10,20,30,40,50,80},
            every x tick label/.append style={font=\tiny},
            every y tick label/.append style={font=\tiny},
            tick pos=left,
            axis y line*=left,
            axis x line*=bottom,
           title={{Spanish \\ (Learner Accuracy)}},
            title style={yshift=-.2cm,font=\tiny,align=center},
            ymin=0.0,ymax=1,
            ylabel shift={-.2cm},
            ylabel near ticks,
            ylabel={Accuracy},
            ylabel style={font=\small},
        ]
        \addplot [style={blue},mark=o,] table[x=Number,y=rule]{\spadata};
        \addplot [style={red},mark=x,] table[x=Number,y=norule]{\spadata};
    \end{axis}
\end{tikzpicture}
\begin{tikzpicture}[trim left=-3.7cm,trim right=0cm]
    \begin{axis}[
            every axis plot post/.style={/pgf/number format/fixed},
            width=4.5cm,
            height=3cm,
            ymajorgrids=false,
            yminorgrids=false,
            legend style={draw=none},
            xtick={5,10,20,30,40,50,80},
            xticklabels={5,10,20,30,40,50,80},
            every x tick label/.append style={font=\tiny},
            every y tick label/.append style={font=\tiny},
            tick pos=left,
            axis y line*=left,
            axis x line*=bottom,
            title={{Spanish \\ (Learner Confidence)}},
            title style={yshift=-.2cm,font=\tiny,align=center},
             ymin=1,ymax=5,
            ylabel style={font=\small},
        ]
        \addplot [style={blue},mark=o] table[x=Number,y=rule]{\spaconfdata};
        \addplot [style={red},mark=x,] table[x=Number,y=norule]{\spaconfdata};
    \end{axis}
\end{tikzpicture}

\begin{tikzpicture}[trim left=-0.85cm,trim right=0cm]
    \begin{axis}[
            every axis plot post/.style={/pgf/number format/fixed},
            width=4.7cm,
            height=3cm,
            ymajorgrids=false,
            yminorgrids=false,
            legend style={draw=none,at={(1.35,1.7)},font=\tiny},
            xtick={5,10,20,30,40,50,120},
            xticklabels={5,10,20,30,40,50,120},
            every x tick label/.append style={font=\tiny},
            every y tick label/.append style={font=\tiny},
            tick pos=left,
            axis y line*=left,
            axis x line*=bottom,
            xmax=121,
             title={{Greek \\ (Learner Accuracy)}},
            title style={yshift=-.2cm,xshift=-0.5cm,font=\tiny,align=center},
            ymin=0.0,ymax=1,
            ylabel shift={-.2cm},
            ylabel near ticks,
            ylabel={Accuracy},
            ylabel style={font=\small},
        ]
         \addplot [style={blue},mark=o,] table[x=Number,y=rule]{\greekdata};
        \addplot [style={red},mark=x,] table[x=Number,y=norule]{\greekdata};
        \legend{with rules, without rules}
    \end{axis}
\end{tikzpicture}
\hspace{-.4cm}
\begin{tikzpicture}[trim left=-3.7cm,trim right=0cm]
    \begin{axis}[
            every axis plot post/.style={/pgf/number format/fixed},
            width=4.7cm,
            height=3cm,
            ymajorgrids=false,
            yminorgrids=false,
            legend style={draw=none},
            xtick={5,10,20,30,40,50,120},
            xticklabels={5,10,20,30,40,50,120},
            every x tick label/.append style={font=\tiny},
            every y tick label/.append style={font=\tiny},
            tick pos=left,
            axis y line*=left,
            axis x line*=bottom,
            xmax=121,
             title={{Greek \\ (Learner Confidence)}},
            title style={yshift=-.2cm,font=\tiny,align=center},
            ymin=1,ymax=5,
            ylabel style={font=\small},
        ]
        \addplot [style={blue},mark=o,] table[x=Number,y=rule]{\greekconfdata};
        \addplot [style={red},mark=x,] table[x=Number,y=norule]{\greekconfdata};
    \end{axis}
\end{tikzpicture}
\vspace{-3mm}
   \caption{ Learner accuracy and confidence in correct answers with and without access to rules against the number of attempted examples ($x$-axis ). Learners achieve higher accuracy with increasing confidence with fewer examples when they have access to rules.} 
    \label{fig:rule_effect}
    \vspace{-1em}
\end{figure}

We fit LME models on our data by varying the number of first $n$ attempted examples $n\!=\![5,10,20,30,40,50,\text{all}]$. Each fitted LME model gives us an intercept which informs us of the learner accuracy in absence of rules, and the fixed-effect coefficient $\mathbf{\beta}$ which informs us about the gain with rules. 
As shown in Figure \ref{fig:rule_effect}, it is clear that learners having access to our automatically extracted rules achieve higher accuracy with fewer examples as compared to without. 
As expected, with an increasing number of attempted examples the gap in accuracy between the two settings reduces.
Interestingly, we find that the rules still have a significant effect on the learner's confidence even later in the learning process. 
This suggests that with our  rules learners require fewer examples to infer the patterns governing each lexical choice and further get more confident in their understanding. 
This is encouraging  as in true settings the learning exercise would be conducted for \emph{every} focus word that the learner is attempting to learn, and because this process will have to be repeated many times, making it more efficient is of significant value.
In Appendix \ref{app:pvalues} we report the $p$-value for the fitted LME models which shows that the positive gains from the presence of rules are most significant for $\leq$20 examples for Spanish and for all examples for Greek. 

Overall, we find our extracted rules help both Spanish and Greek learners in their learning process. 
We note that the results on Greek are promising as it does not enjoy the same luxuries as Spanish in having a high-quality lemmatizer or word aligner.
This is encouraging especially for researchers involved in the revival efforts of endangered languages.

\begin{figure}

\begin{tikzpicture}[trim left=-0.85cm,trim right=0cm]
\def\MarkSize{1.2pt}
  \protected\def\ToWest#1{%
    \llap{#1\kern\MarkSize}\phantom{#1}%
  }
  \protected\def\ToSouth#1{%
    \sbox0{#1}%
    \smash{%
      \rlap{%
        \kern-.5\dimexpr\wd0 + \MarkSize\relax
        \lower\dimexpr.375em+\ht0\relax\copy0 %
      }%
    }%
    \hphantom{#1}%
  }
    \begin{axis}[
            width=4.4cm,
            height=3.0cm,
            ymajorgrids=false,
            yminorgrids=false,
            legend style={draw=none},
            every node near coord/.append style={font=\tiny},
            every x tick label/.append style={font=\tiny},
            every y tick label/.append style={font=\tiny},
            tick pos=left,
            axis y line*=left,
            axis x line*=bottom,
             title={\textbf{Spanish}},
            title style={yshift=-.4cm,font=\tiny,align=center},
            ymin=0.0,ymax=0.35,
            ylabel shift={-.2cm},
            ylabel near ticks,
            ylabel={{Rule Effect ($\beta$)}},
            ylabel style={font=\small},
            ]
    \addplot[
        scatter/classes={a={blue}, b={red}},
        scatter, mark=*, only marks, 
        scatter src=explicit symbolic,
        nodes near coords*={\Label},
        mark size=1.2pt,
        visualization depends on={value \thisrow{label} \as \Label} 
    ] table [meta=class] {
        x y label class
        0.925	-0.0619	figure	a
        0.6	0.3174	oil	a
        0.883	-0.04583	wave	a
        0.85	0.1	language	a
        0.85	0.025	pill	a
        0.89	0.06	ticket	a
        0.483	0.1	wall	a
        0.75	0.2	vote	a
        0.6125	0.20309	\ToSouth{farmer}	a
    };
    \addplot [red, mark=none ] table[y={create col/linear regression={y=y}}] {
        x y label class
        0.925	0.0619	figure	a
        0.6	0.3174	oil	a
        0.883	-0.04583	wave	a
        0.85	0.1	language	a
        0.85	0.025	pill	a
        0.89	0.06	ticket	a
        0.483	0.1	wall	a
        0.75	0.2	vote	a
        0.6125	0.20309	farmer	a
    };
\end{axis}
\end{tikzpicture}
\hspace{-.2cm}
\begin{tikzpicture}[trim left=-3.7cm,trim right=0cm]
    \begin{axis}[
            width=4.4cm,
            height=3.0cm,
            ymajorgrids=false,
            yminorgrids=false,
            legend style={draw=none},
            every node near coord/.append style={font=\tiny},
            every x tick label/.append style={font=\tiny},
            every y tick label/.append style={font=\tiny},
            tick pos=left,
            axis y line*=left,
            axis x line*=bottom,
             title={\textbf{Greek}},
            title style={yshift=-.4cm,font=\tiny,align=center},
            ymin=0.0,ymax=0.35,
            ylabel shift={-.2cm},
            ylabel near ticks,
            ylabel style={font=\small},
            ]
    \addplot[
        scatter/classes={a={blue}, b={red}},
        scatter, mark=*, only marks, 
        scatter src=explicit symbolic,
        nodes near coords*={\Label},
        mark size=1.2pt,
        visualization depends on={value \thisrow{label} \as \Label} 
    ] table [meta=class] {
        x y label class
        0.65	0.2739	tour	a
        0.7	0.18504	turn	a
        0.8875	0.1025	bill	a
        0.6375	0.04946	roof	a
        0.59	0.1814	break	a
        0.65	0.02	old	a
        0.85	0.05	tie	a
        0.7375	0.05	effect	a
        0.51	0.115	wheel	a
        0.75	0.13527	bone	a
    };
    \addplot [red, mark=none] table[y={create col/linear regression={y=y}}] {
        x y label class
        0.65	0.2739	tour	a
        0.7	0.18504	turn	a
        0.8875	0.1025	bill	a
        0.6375	0.04946	roof	a
        0.59	0.1814	break	a
        0.65	0.02	old	a
        0.85	0.05	tie	a
        0.7375	0.05	effect	a
        0.51	0.115	wheel	a
        0.75	0.13527	bone	a
    };
\end{axis}
\end{tikzpicture}
\vspace{-3mm}
   \caption{Rules help more for words where learners do worse. x-axis is the (avg.) learner accuracy (without rules) for first 20 examples.} 
    \label{fig:word_effect}
    \vspace{-1em}
\end{figure}
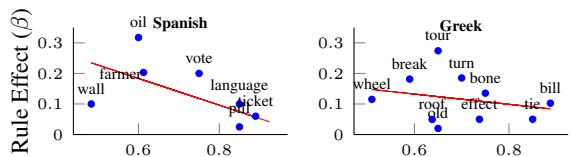

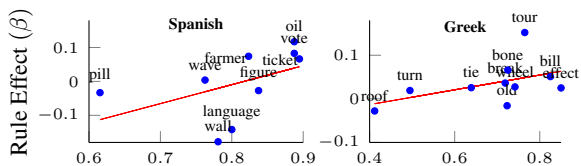
\begin{figure}

\begin{tikzpicture}[trim left=-0.95cm,trim right=0cm]
\def\MarkSize{1.2pt}
  \protected\def\ToWest#1{%
    \llap{#1\kern\MarkSize}\phantom{#1}%
  }
  \protected\def\ToSouth#1{%
    \sbox0{#1}%
    \smash{%
      \rlap{%
        \kern-.5\dimexpr\wd0 + \MarkSize\relax
        \lower\dimexpr.375em+\ht0\relax\copy0 %
      }%
    }%
    \hphantom{#1}%
  }
    \begin{axis}[
            width=4.4cm,
            height=3.2cm,
            ymajorgrids=false,
            yminorgrids=false,
            legend style={draw=none},
            every node near coord/.append style={font=\tiny},
            every x tick label/.append style={font=\tiny},
            every y tick label/.append style={font=\tiny},
            tick pos=left,
            axis y line*=left,
            axis x line*=bottom,
            title={\textbf{Spanish}},
            title style={yshift=-.5cm,font=\tiny,align=center},
            ymin=-0.18,ymax=0.18,
            xmin = 0.60, xmax=0.9,
            ylabel shift={-.2cm},
            ylabel near ticks,
            ylabel={Rule Effect $(\beta)$},
            ylabel style={font=\small},
            ]
    \addplot[
        scatter/classes={a={blue}, b={red}},
        scatter, mark=*, only marks, 
        scatter src=explicit symbolic,
        nodes near coords*={\Label},
        mark size=1.2pt,
        visualization depends on={value \thisrow{label} \as \Label} 
    ] table [meta=class] {
        x y label class
       0.8375	-2.72E-02	figure	a
        0.8875	1.17E-01	oil	a
        0.7625	3.94E-03	wave	a
        0.8	-1.42E-01	language	a
        0.6154	-3.32E-02	pill	a
        0.8947	6.63E-02	\ToSouth{ticket}	a
        0.7808	-1.78E-01	wall	a
        0.8875	8.27E-02	vote	a
        0.8235	7.41E-02	\ToSouth{farmer}	a
    };
    \addplot [red, mark=none] table[y={create col/linear regression={y=y}}] {
        x y label class
        0.8375	-2.72E-02	figure	a
        0.8875	1.17E-01	oil	a
        0.7625	3.94E-03	wave	a
        0.8	-1.42E-01	language	a
        0.6154	-3.32E-02	pill	a
        0.8947	6.63E-02	ticket	a
        0.7808	-1.78E-01	wall	a
        0.8875	8.27E-02	vote	a
        0.8235	7.41E-02	farmer	a
    };
\end{axis}
\end{tikzpicture}
\hspace{-.2cm}
\begin{tikzpicture}[trim left=-3.7cm,trim right=0cm]
    \begin{axis}[
            width=4.1cm,
            height=3.2cm,
            ymajorgrids=false,
            yminorgrids=false,
            legend style={draw=none},
            every node near coord/.append style={font=\tiny},
            every x tick label/.append style={font=\tiny},
            every y tick label/.append style={font=\tiny},
            tick pos=left,
            axis y line*=left,
            axis x line*=bottom,
             title={\textbf{Greek}},
            title style={yshift=-.5cm,font=\tiny,align=center},
            ymin=-0.1,ymax=0.18,
            xmin = 0.4, xmax=0.85,
            ylabel shift={-.2cm},
            ylabel near ticks,
            ylabel style={font=\small},
            ]
    \addplot[
        scatter/classes={a={blue}, b={red}},
        scatter, mark=*, only marks, 
        scatter src=explicit symbolic,
        nodes near coords*={\Label},
        mark size=1.2pt,
        visualization depends on={value \thisrow{label} \as \Label} 
    ] table [meta=class] {
        x y label class
        0.7647	0.15212	tour	a
        0.4948	0.01916	turn	a
        0.825	0.050974	bill	a
        0.4118	-0.02784	roof	a
        0.7188	0.03621	break	a
        0.7232	-0.01582	old	a
        0.6389	0.02553	tie	a
        0.85	2.50E-02	effect	a
        0.7417	0.02737	wheel	a
        0.725	0.06592	bone	a
    };
    \addplot [red, mark=none] table[y={create col/linear regression={y=y}}] {
        x y label class
        0.7647	0.15212	tour	a
        0.4948	0.01916	turn	a
        0.825	0.050974	bill	a
        0.4118	-0.02784	roof	a
        0.7188	0.03621	break	a
        0.7232	-0.01582	old	a
        0.6389	0.02553	tie	a
        0.85	2.50E-02	effect	a
        0.7417	0.02737	wheel	a
        0.725	0.06592	bone	a
    };
\end{axis}
\end{tikzpicture}
\vspace{-3mm}
   \caption{Rules help more for words where the model performs well. x-axis is model accuracy per word.} 
    \label{fig:word_effect_model}
    \vspace{-1em}
\end{figure}
\paragraph{Do the extracted rules result in  increased learner confidence? }
While answering the questions we ask the learner to mark how confident they are in their answer. As before, we fit LME models for each $n$ using annotator confidence as the response variable $\mathbf{Y}$ and presence of rules as the fixed-effect. We find that the learners' confidence in the correct answer increases more when they are provided rules (Figure~\ref{fig:rule_effect}) for both the languages.

\paragraph{Do the extracted rules help some words \emph{more} over others? }
Since the focus words may vary in the difficulty level, we check if our extracted rules are more effective for some words over others. So, we fit a LME model on each  focus word and compute the $\mathbf{\beta}$ coefficient to measure the effect of rules on learner accuracy after 20 attempted examples.\footnote{Because analysis revealed that rules are more effective earlier in the learning process.} 
We plot the $\mathbf{\beta}$ coefficient with the accuracy (averaged across all learners) for each focus word when they didn't have rules in Figure \ref{fig:word_effect} and find that words on which the learners performed the worst such as \emph{wall, oil, farmer,} and  \textit{vote} for Spanish,   benefit most by our rules. 
Similar observations can be seen for Greek where  learners are benefited more for words (\emph{break, wheel, tour, old, roof}) on which they performed the worst. 
Some of these words, in fact, indeed have  finer semantic subdivisions than the rest.
For instance, the choices for \emph{farmer}: \emph{agricultor} refers to exclusively the one who works the land, harvests, sows, etc., whereas \emph{granjero} is less formal referring to the one who manages a farm, or works or lives on it.\footnote{This is based on explanations collected from native Spanish speakers, which can also be found in Appendix \ref{app:pvalues}} This analysis shows that, encouragingly, our rules are especially helping learners with more difficult words.

We also plot the $\mathbf{\beta}$ coefficient with the lexical model accuracy (Figure \ref{fig:word_effect_model}) and find a positive correlation, meaning that rules help more for words where the model performs well.
This  suggests that if we can develop more accurate models with an equal level of interpretability, the learning effect might become even stronger.

\section{Related Work}
\paragraph{Computer-assisted language learning} CALL systems have been increasingly using NLP  for creating learning content. Both SMILLE \cite{zilio-etal-2017-using}  and WERTi \cite{meurers-etal-2010-enhancing} aim to help the text understanding process  by highlighting linguistic structures using hand-written rules and automatically acquired syntactic analysis. \emph{Apertium}~\cite{tyers-etal-2012-flexible}, a rule-based MT system, while not aimed at language learning, does use human- and machine-readable rules, whose formalism can account for only fixed-length ordered contexts restricting their application.
Further, these rules use a combination of only lemma and POS tags while our framework uses more features. 

\paragraph{Cross-lingual word sense disambiguation} CL-WSD disambiguates a word in-context by providing appropriate translation across languages. \citet{lefever-hoste-2010-semeval} construct a dataset (25 ambiguous English nouns across six languages) semi-automatically from parallel corpora  which are then verified by expert translators. Such lexical choice tasks have been created also for evaluating MT systems \cite{rios-gonzales-etal-2017-improving, rios-etal-2018-word}. However, these methods cover a limited set of words (mostly nouns) and require some manual intervention during the data creation process. To the best of our knowledge, our proposed pipeline is the only fully automated one that extracts several ambiguous words across multiple POS tags.

\section{Future Work}
While we have  demonstrated the efficacy of our extracted rules  in teaching new words for two languages, we  plan to apply our framework on much less-resourced languages which  have fewer available learning resources where  learners would benefit more from an automated system.
We also plan to use automated methods such as selection using model confidence to select `representative' examples for the learning setup instead of using the native speakers. Furthermore, multimodal features have proven their utility in automatic methods for lexical acquisition \cite{hewitt-etal-2018-learning}, and we plan to examine their effectiveness for L2 learning.
 

\section*{Acknowledgements}
The authors are grateful to the anonymous reviewers who took the time to provide many interesting
comments that made the paper significantly better. 
We would also like to thank  Nikolai Vogler for the original interface for data annotation, and all the learners  for their participation in our study, and without whom this study would not have been possible or meaningful.
This work is sponsored by the Waibel Presidential Fellowship and by the
National Science Foundation under grants 1761548 and 2125466.

\bibliographystyle{acl_natbib}
\bibliography{emnlp}

\clearpage
\newpage

\appendix
\newpage
\appendix

\section{Automated Evaluation}
\label{app:auto}
\subsection{Identifying Semantic Subdivisions}
\begin{figure*}%
\centering
\subfigure[Spanish]{
\label{fig:spanish}%
\includegraphics[width=0.5\textwidth]{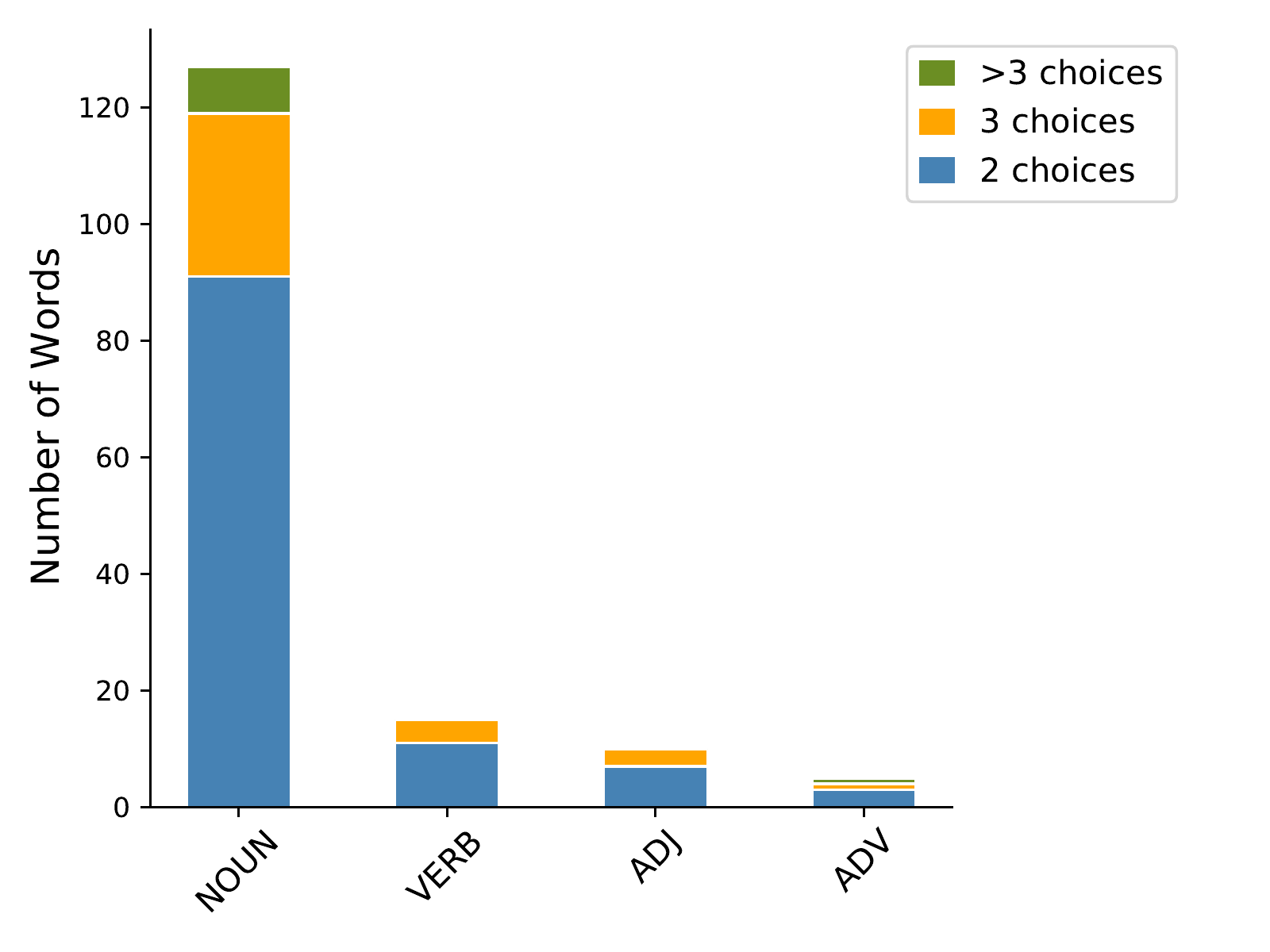}}
~
\subfigure[Greek]{
\label{fig:greek}%
\includegraphics[width=0.5\textwidth]{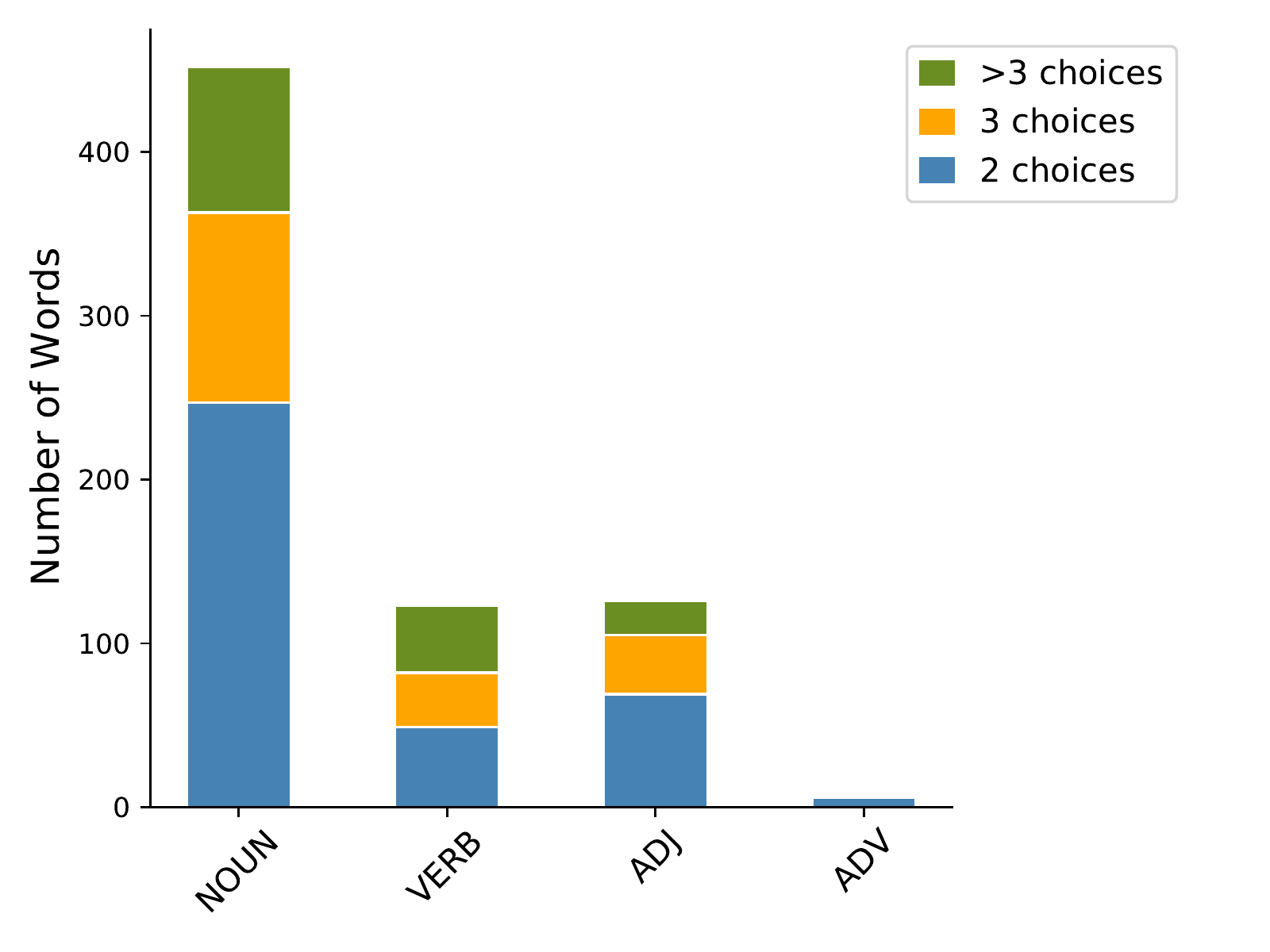}
}
  \caption{Distribution of number of lexical choices for each POS tag.}%
\label{fig:pos}%
\end{figure*}

In Section 3, we describe the procedure for identifying focus words in L1. For the step of \textit{filter on entropy} within that procedure, we use a threshold of $0.69$ so focus words having an entorpy $H>0.69$ are selected in that step. For binary lexical choices, an ambiguous word would be aligned to each choice with uniform distribution and the entropy in that case would be $0.69$. Hence we are interested in words that  exceed this minimum threshold.
In Figure \ref{fig:pos} we show the distribution of number of lexical choices for all extracted focus words, filtered by each POS tag for Spanish and Greek. 
We check the CEFR levels\footnote{\url{https://en.wikipedia.org/wiki/Common_European_Framework_of_Reference_for_Languages}} which  measure  the reading proficiency in a language. We use the automated tool provided by Duolingo\footnote{\url{https://cefr.duolingo.com/}} (currently available only for Spanish and English) to get these levels and find that 60\% of the extracted Spanish lexical choices belong to the B level which is the intermediate level and 20\% belong to the advanced level.
This suggests that the identified words are indeed more challenging.

\subsection{Model Hyperparameters and Results}
\label{app:setup}
For the LinearSVM and DTree models, we clean the data to remove punctuation and extract features within a 3-word window of the focus word.
As mentioned before, we train a lexical selection model for each focus word and in Table \ref{tab:results}, \ref{tab:results-1}, \ref{tab:results-2}, \ref{tab:results-3} we report the individual accuracy for the test accuracy for LinearSVM, DTree, BERT and the baseline method across both Spanish and Greek. We also provide the train accuracy for our main model, LinearSVM. 
\paragraph{LinearSVM}

We perform a grid search over the following hyperparameters: \texttt{C} $=$ [0.001, 0.01], \texttt{class weight} $=$['balanced', None].

\paragraph{DTree}
We also experimented with other  interpretable models such as decision trees \cite{quinlan1986induction} using the CART algorithm \cite{breiman1984classification}, however we found them to be performing worse than the SVM model.
We used the following hyperparameters:  \texttt{criterion} $=$ [gini, entropy],  \texttt{max depth} $=$ [6,15], \texttt{min impurity decrease} $= 1e^{-3}$.

\paragraph{BERT}
We compare the interpretable models LinearSVM and DTree with more complex neural model based on the popular BERT \citep{devlin2018bert}.
We retain the same hyperparameters as the original paper using 768 dimentions for the encoder representations. 
We train the model for  20 epochs, using the AdamW optimzer with a learning rate of $5e-5$.

\begin{table*}[t]
    \begin{center}
    \resizebox{\textwidth}{!}{
    \begin{tabular}{c|l}
    \toprule
    \textbf{Lexical Choice} & \textbf{feature $\longrightarrow \langle$ rule name $\rangle$  $\langle$ feature value$\rangle$ }  \\
    \midrule
    muro & Bigram $\longrightarrow$ Short phrases:   ('climb', 'wall'), ('city', 'wall'), ('brick', 'wall')   \\
         & Lemma $\longrightarrow$ Words: break, climb   \\
         & WSD $\longrightarrow$ Concepts: `city' as in a large and densely populated urban area (city.n.01) \\
    \midrule
    pared & Bigram $\longrightarrow$ Short phrases:   ('face', 'wall'), ('hang', 'wall'), ('picture', 'wall')   \\
         & Lemma $\longrightarrow$ Words:  ear, hang, room   \\
    \bottomrule
    \end{tabular}
    }
    \caption{Human-readable rules extracted for the ambiguous word \textit{wall} (top-6 rules per lexical choice). }
   \label{tab:wall}
     \end{center}
\end{table*}

\section{Human Evaluation}
\subsection{Rule Templates}
\label{app:templates}
The human-understandable “rules” are essentially those features from the training set which the model thought were important for determining the correct label (i.e. features that were given higher weights for a given label). 
In particular, for each label (i.e. \emph{muro} or \emph{pared}), we choose the top-20 features. We then group these individual features together by their feature types, for instance, all the bigram features are grouped under the category called ``Short Phrase'', lemma features are grouped under the category ``Words'', and the WSD features are first expanded into their natural language form using the WordNet \cite{miller1995wordnet} knowledge base and then grouped under the category ``Concepts'', as shown in Table \ref{tab:wall}.

\subsection{Language Learning Interface}
\label{app:interface}
\begin{figure*}%
\centering
\subfigure[Rules for ``pared'' vs ``muro''.]{
\label{fig:rules}%
\includegraphics[width=\textwidth]{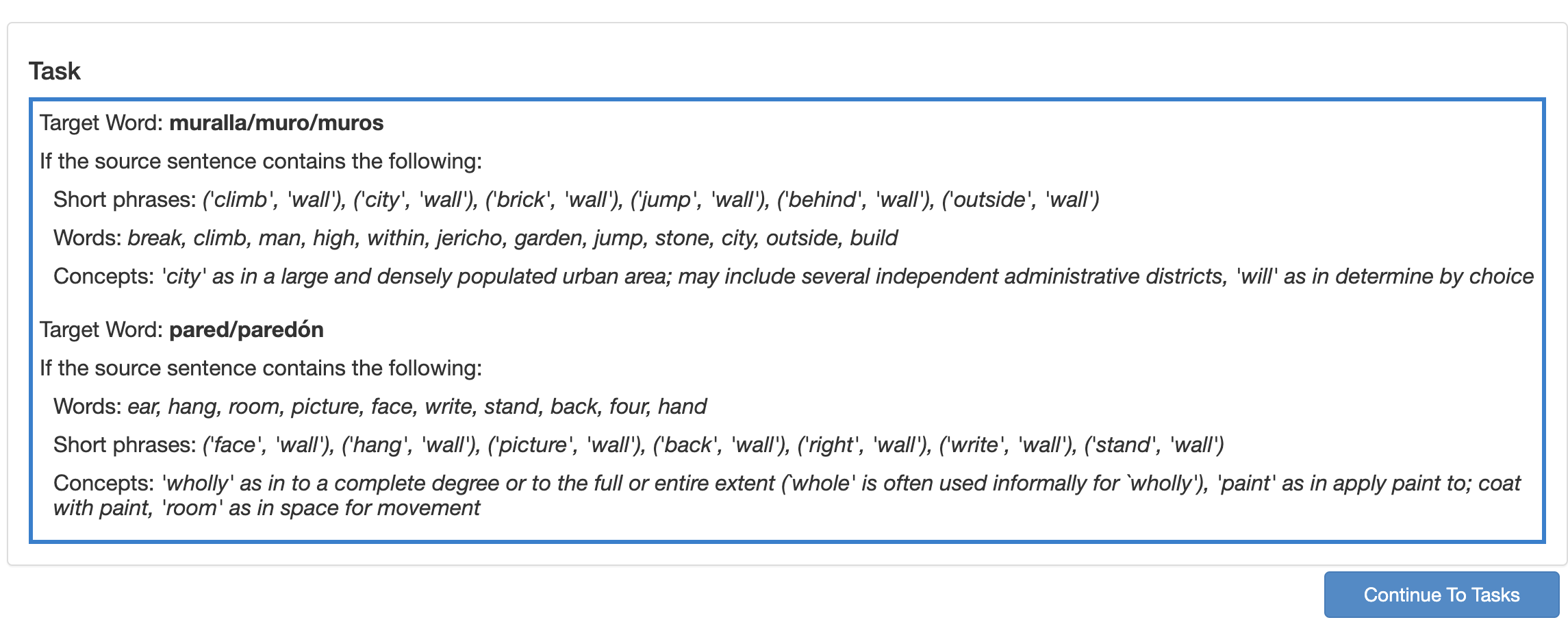}}

\subfigure[Rules for the correct answer are displayed  to the learner after each question. Individual rules which apply to the given example are highlighted for the convenience of the learner.]{
\label{fig:task}%
\includegraphics[width=\textwidth]{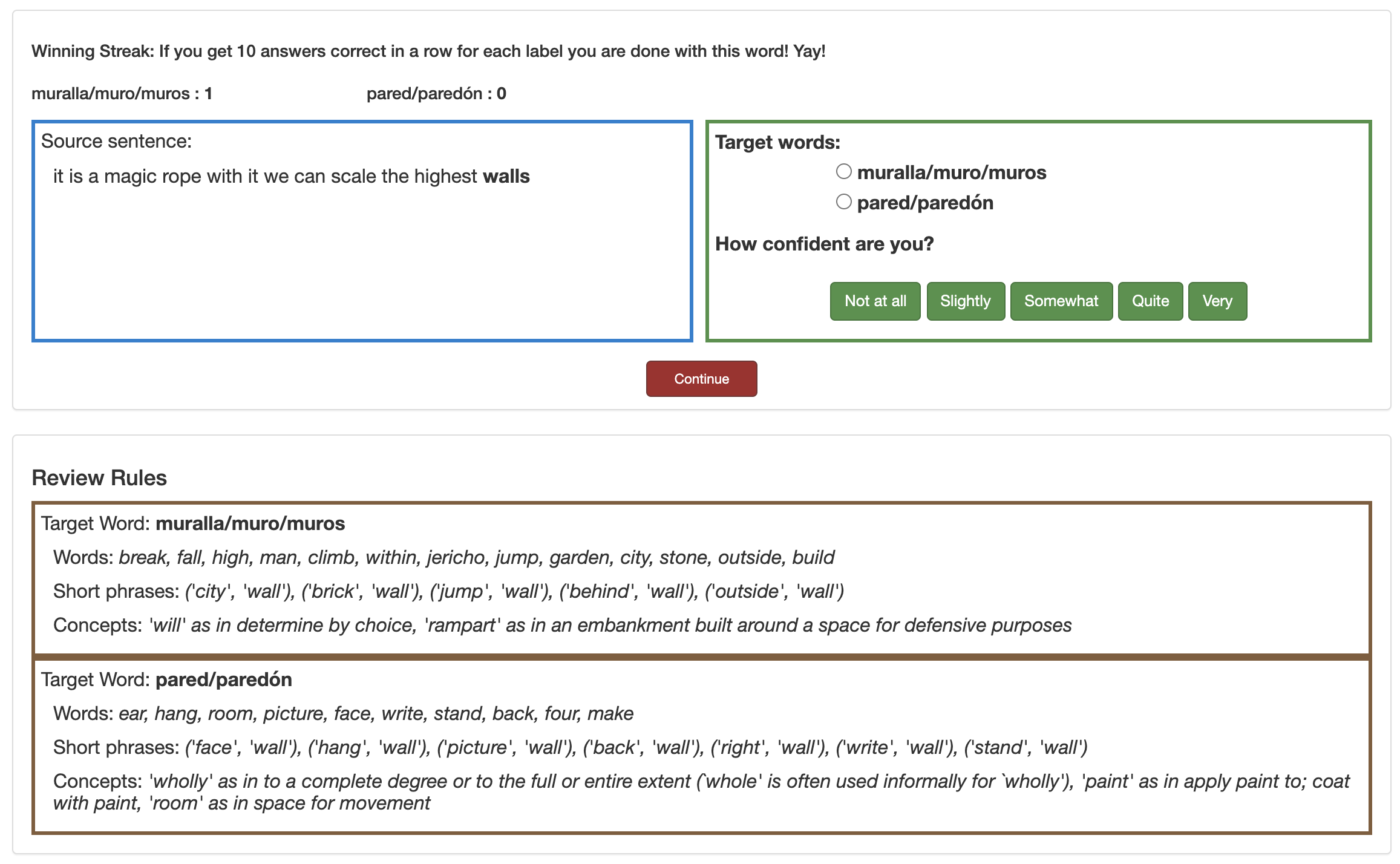}
}
  \caption{User interface for human language learning experiment. }%
\label{fig:interface}%
\end{figure*}

In our proposed language learning setup, the learner is first presented with a screen showing concise explanations for each lexical choice (Figure \ref{fig:rules}). They can take as much time as they require for reviewing the rules and then proceed to the tasks. Within each task, the learner is then shown an English sentence with the focus word highlighted and a set of possible lexical choices. The page also displays the concise explanations for the learner to refer if they wish to (Figure \ref{fig:task}). The learner is required to select one of the lexical choices and mark how confident they are in their answer. Once submitted, the learner is immediately shown the correct answer along with individual rules that applied to that example highlighted (Figure 2 in main text). Learners took (avg.) 3-4 hours in total to complete all tasks.
Table \ref{tab:data} presents the tasks performed by the respective Spanish and Greek learners.
Since  English speakers might not be familiar with the Greek alphabet, we display the  English transliteration of the respective Greek words. 
Some of the lexical choices (e.g. \textit{muralla/muro/muros}) contain multiple inflections of the same lemma (\textit{muro}). 
This is due to errors in the automatic Spanish lemmatizer which failed to correctly map the inflections to a single lemma. 
We therefore run an edit-distance based post-processing to combine lexical choices having the same prefix. We note that this simple heuristic might not be ideal for languages such as Indonesian that use affixes and/or reduplication with far-from-perfect lemmatizers; nevertheless such a post-processing method, when applied carefully, helps fix many of the erroneous lemmatization issues.

\subsection{Representative Example Selection}
\label{app:lexical}
The shortlisted words for both the Spanish and Greek study can be found in Table \ref{tab:data}.
We use  native speakers to filter sentences that have sufficient context for correctly identifying a lexical choice. We enlist 3 Spanish native speakers who each annotate roughly 200 examples each for 10 English focus words. 
The inter-annotator agreement for Spanish, computed using Fleiss' kappa is 0.77.
For Greek, we use 2 native speakers to annotate 10 English words. For 7 out of 10 words we did not always have access to 2 native speakers so we relied on  a single expert annotator. 
The (avg.) inter-annotator agreement for the remaining 3 words (\emph{tour, tie, bill}) between the two annotators is 0.83.
Of the 10 selected words, we discard words/lexical choices which have $<10$ examples  on which all native speakers agree (Table \ref{tab:data}) giving us 9 English words for the Spanish study and 10 English for the Greek study.

\begin{table*}[t]
\small
    \centering
    \resizebox{\textwidth}{!}{
    \begin{tabular}{c|c||c|c}
    \toprule
    \multicolumn{2}{c}{\textbf{Spanish}} & \multicolumn{2}{c}{\textbf{Greek}} \\\toprule
 (en) word & (es) lexical choices & (en) word & (el) lexical choices \\
    \midrule
    \multirow{2}{*}{wall.\textit{N}} & muralla/muro/muros: 33, pared/paredón: 60 & \multirow{2}{*}{bill.\textit{N}} & \textgreek{χαρτονόμισμα: 40, λογαριασμός: 40, νόμος/νομοσχέδιο: 40}\\
    & & &  (charton\'omisma, logariasm\'os, n\'omos/nomosch\'edio) \\\midrule
    
    \multirow{2}{*}{farmer.\textit{N}} & agricultor: 29, granjero: 48   & \multirow{2}{*}{tour.\textit{N}} & \textgreek{θητεία:23, περιοδεία: 29, ξενάγηση: 33 }\\
    & & &  (thit\'ia, periode\'ia, xen\'agisi) \\  \midrule
    
    \multirow{2}{*}{figure.\textit{N}} & cifra/cifras: 87, figura: 85 & \multirow{2}{*}{break.\textit{JJ}} & \textgreek{σπάω: 40, ράγομαι: 40, ξεσπάω: 40, διαρρηγνύω: 40}\\
    & & &  (sp\'ao, r\'agomai, ksesp\'ao, diarrign\'io) \\  \midrule
    
    \multirow{2}{*}{vote.\textit{N}} & votemos/voto: 77, votación: 75 & 
    \multirow{2}{*}{turn.\textit{JJ}} & \textgreek{στρίβω: 40, χαμηλώνω: 40, απορρίπτω: 40, καταδίδω: 40, σβήνω: 34}\\
    & & & (strívo,chamilòno, aporrípto,katadído, svíno)\\  \midrule
    
    \multirow{2}{*}{oil.\textit{N}} & aceite: 81, óleo/petróleo/petrolera/petrolero: 74 &
    \multirow{2}{*}{roof.\textit{N}} & \textgreek{ταράτσα: 40,οροφή: 40, στέγη: 39 }\\
    & & & (tar\'atsa, orof\'i, st\'egi) \\\midrule
    
    \multirow{2}{*}{wave.\textit{N}} & onda: 55, ola: 40, \textit{\textcolor{red}{oleado: 0}} & 
    \multirow{2}{*}{wheel.\textit{N}} & \textgreek{τροχός: 40, ρόδα: 40, τιμόν: 40}\\
    & & & (troh\'os, r\'oda, tim\'oni)\\  \midrule
    
    \multirow{2}{*}{pill.\textit{N}} & pastilla: 41, somn\'ifero: 27, \textit{\textcolor{red}{p\'ildora: 3}} & 
    \multirow{2}{*}{old.\textit{JJ}} & \textgreek{αρχαίος: 40, κλασικ: 21, έτος: 40, ηλικιωμένος: 40, παραδοσιακός: 36} \\
    & & & (archaios, klasikos, etos, elikiomenos,paradosiakos)\\ \midrule
    
    \multirow{2}{*}{language.\textit{N}} & idioma: 52, lenguaje: 68 & 
    \multirow{2}{*}{turn.\textit{JJ}} & \textgreek{στρίβω: 40, χαμηλώνω: 40, απορρίπτω: 40, καταδίδω: 40, σβήνω: 34} \\
    & & & (strívo, chamil\'ono, aporr\'ipto, katad\'ido,svíno)\\ \midrule

    \multirow{2}{*}{ticket.\textit{N}} & multa: 24, boleto: 23, \textit{\textcolor{red}{pasaje: 0}} &
    \multirow{2}{*}{effect.\textit{N}} & \textgreek{παρενέργεια: 40, επίδραση: 40, εφέ: 40}\\
    & & & (paren\'irgeia, ep\'idrasi, ef\'e) \\  \midrule
    
    \multirow{2}{*}{\textit{\textcolor{red}{servant.\textit{N}}}} & sirvienta/sirviente: 39, \textit{\textcolor{red}{servidor/servidora: 8}}, \textit{\textcolor{red}{siervo/siervos: 10}} &
    \multirow{2}{*}{bone.\textit{N}} & \textgreek{μυελός: 40, οστό: 40, Μπόουν: 40}\\
    & & & (myel\'os,  ost\'o, bone) \\
    \bottomrule
    \end{tabular}
    }
    \caption{Example tasks with their lexical choices selected for Spanish and Greek learning setup. Words/choices marked in \textit{\textcolor{red}{red}} are discarded from the language learning setup as they have $\leq10$ filtered examples from the represenative example selection step.}
    \label{tab:data}
    \vspace{-1em}
\end{table*}

\subsection{Results}
\label{app:pvalues}
In Table \ref{tab:lmm} we report the p-values for the linear mixed effect (LME) models fitted on predicting learner accuracy with rules as fixed-effect. The results show that the positive effect of rules on accuracy is statistical significant up to first 20 attempted examples for Spanish and up to all examples for Greek.
\begin{table}[ht]
    \begin{center}
    \resizebox{\columnwidth}{!}{
    \begin{tabular}{c|c|c|c}
    \toprule
\textbf{Number} & \textbf{Fixed-effect coefficient ($\mathbf{\beta})$} & \textbf{Spanish p-value}  & \textbf{Greek p-value} \\
\midrule
5 &0.118 &	0.013\textbf{**} & \textbf{$4.50e^{-09}$***} \\
10 & 0.112 &	0.009\textbf{***} & \textbf{$1.64e^{-07}$***}\\
20	& 0.056 &	0.070\textbf{*} & \textbf{$1.32e^{-06}$***} \\
30	& 0.039 &	0.131 & \textbf{$4.23e^{-05}$***}\\
40	& 0.017 &	0.462 &  \textbf{$7.22e^{-05}$***}\\
50	& 0.007 &	0.718 & \textbf{$0.00015$***}\\
\midrule
All	& 0.006 &	0.739 & \textbf{$0.00173$**}\\
\bottomrule
\end{tabular}
    }
    \caption{$p$-value tests show that the fixed-effect of presence of rules for predicting learner accuracy is statistical significant up to first 20 attempted examples for Spanish and up to all examples for Greek. Significance codes: ‘***’: 0.01, ‘**’: 0.05, ‘*’: 0.1.}
   \label{tab:lmm}
     \end{center}
\end{table}

\begin{table*}[t]
    \begin{center}
    \resizebox{\textwidth}{!}{
    \begin{tabular}{l|c|l|c|l|c}
Focus word & FreqBaseline - DTree - LinearSVM Train/Test - BERT & Focus word & FreqBaseline - DTree - LinearSVM Train/Test - BERT &Focus word & FreqBaseline - DTree - LinearSVM Train/Test - BERT \\
\midrule
specifically.\textit{RB} & 0.67 - 0.7 - 0.67 / 0.67 - 0.72 & block.\textit{N} & 0.58 - 0.69 - 0.92 / 0.79 - 0.86 & desert.\textit{V} & 0.53 - 0.55 - 0.96 / 0.73 - 0.95 \\ 

transfer.\textit{N} & 0.64 - 0.71 - 0.92 / 0.69 - 0.72 & slipper.\textit{N} & 0.7 - 0.7 - 0.7 / 0.7 - 0.63 & mushroom.\textit{N} & 0.42 - 0.39 - 0.91 / 0.44 - 0.49 \\ 

pen.\textit{N} & 0.73 - 0.74 - 0.73 / 0.73 - 0.73 & foundation.\textit{N} & 0.43 - 0.5 - 0.96 / 0.64 - 0.69 & mercy.\textit{N} & 0.58 - 0.67 - 0.83 / 0.72 - 0.77 \\ 

fry.\textit{V} & 0.57 - 0.71 - 0.84 / 0.69 - 0.88 & bug.\textit{N} & 0.57 - 0.6 - 0.82 / 0.6 - 0.48 & toast.\textit{N} & 0.64 - 0.72 - 0.96 / 0.87 - 0.94 \\ 

cord.\textit{N} & 0.52 - 1.0 - 0.98 / 1.0 - 1.0 & waste.\textit{N} & 0.48 - 0.53 - 0.87 / 0.74 - 0.8 & opponent.\textit{N} & 0.66 - 0.66 - 0.67 / 0.66 - 0.59 \\ 

figure.\textit{N} & 0.53 - 0.55 - 0.93 / 0.76 - 0.93 & hood.\textit{N} & 0.55 - 0.61 - 0.94 / 0.74 - 0.92 & steak.\textit{N} & 0.66 - 0.34 - 0.66 / 0.66 - 0.61 \\ 

poker.\textit{N} & 0.62 - 0.62 - 0.62 / 0.62 - 0.54 & heel.\textit{N} & 0.53 - 0.72 - 0.9 / 0.71 - 0.85 & plot.\textit{N} & 0.6 - 0.6 - 0.99 / 0.76 - 0.88 \\ 

plumber.\textit{N} & 0.57 - 0.57 - 0.57 / 0.57 - 0.65 & replacement.\textit{N} & 0.6 - 0.6 - 0.61 / 0.6 - 0.67 & thick.\textit{JJ} & 0.59 - 0.73 - 0.98 / 0.73 - 0.69 \\ 

pee.\textit{V} & 0.61 - 0.61 - 0.62 / 0.61 - 0.58 & greedy.\textit{JJ} & 0.68 - 0.68 - 0.68 / 0.68 - 0.75 & barber.\textit{N} & 0.75 - 0.69 - 0.76 / 0.75 - 0.74 \\ 

puppet.\textit{N} & 0.53 - 0.53 - 0.9 / 0.51 - 0.46 & basket.\textit{N} & 0.66 - 0.66 - 0.69 / 0.66 - 0.66 & marble.\textit{N} & 0.51 - 0.71 - 0.98 / 0.8 - 0.93 \\ 

bowl.\textit{N} & 0.6 - 0.6 - 0.96 / 0.72 - 0.72 & dump.\textit{N} & 0.61 - 0.64 - 0.9 / 0.64 - 0.67 & lipstick.\textit{N} & 0.51 - 0.49 - 0.93 / 0.4 - 0.53 \\ 

appeal.\textit{N} & 0.75 - 0.8 - 0.95 / 0.82 - 0.93 & promote.\textit{V} & 0.74 - 0.73 - 0.74 / 0.74 - 0.64 & brush.\textit{N} & 0.55 - 0.58 - 0.96 / 0.66 - 0.87 \\ 

fan.\textit{N} & 0.31 - 0.3 - 0.89 / 0.43 - 0.59 & disappoint.\textit{V} & 0.68 - 0.68 - 0.68 / 0.68 - 0.74 & persuade.\textit{V} & 0.7 - 0.7 - 0.74 / 0.7 - 0.58 \\ 

mob.\textit{N} & 0.7 - 0.74 - 0.98 / 0.74 - 0.93 & romance.\textit{N} & 0.58 - 0.64 - 0.98 / 0.73 - 0.79 & raincoat.\textit{N} & 0.65 - 0.65 - 0.64 / 0.65 - 0.56 \\ 

properly.\textit{RB} & 0.42 - 0.42 - 0.42 / 0.42 - 0.41 & jungle.\textit{N} & 0.55 - 0.55 - 0.77 / 0.56 - 0.53 & nail.\textit{N} & 0.6 - 0.67 - 0.93 / 0.81 - 0.9 \\ 

hobby.\textit{N} & 0.58 - 0.58 - 0.68 / 0.58 - 0.63 & pupil.\textit{N} & 0.53 - 0.56 - 0.96 / 0.67 - 0.79 & regularly.\textit{RB} & 0.64 - 0.66 - 0.96 / 0.7 - 0.5 \\ 

stew.\textit{N} & 0.59 - 0.41 - 0.6 / 0.59 - 0.62 & farmer.\textit{N} & 0.66 - 0.7 - 0.93 / 0.77 - 0.84 & pipe.\textit{N} & 0.5 - 0.52 - 0.93 / 0.62 - 0.83 \\ 

bait.\textit{N} & 0.54 - 0.62 - 0.58 / 0.59 - 0.5 & eve.\textit{N} & 0.96 - 0.96 - 0.99 / 0.96 - 0.96 & transitional.\textit{JJ} & 0.56 - 0.56 - 0.9 / 0.56 - 0.52 \\ 

trunk.\textit{N} & 0.74 - 0.76 - 0.77 / 0.74 - 0.81 & oil.\textit{N} & 0.55 - 0.63 - 0.95 / 0.85 - 0.89 & ancestor.\textit{N} & 0.56 - 0.56 - 0.95 / 0.55 - 0.58 \\ 

port.\textit{N} & 0.41 - 0.61 - 0.98 / 0.79 - 0.96 & lock.\textit{N} & 0.8 - 0.85 - 0.91 / 0.85 - 0.89 & cripple.\textit{N} & 0.63 - 0.63 - 0.63 / 0.63 - 0.63 \\ 

shell.\textit{N} & 0.41 - 0.41 - 0.95 / 0.59 - 0.55 & servant.\textit{N} & 0.64 - 0.7 - 0.78 / 0.71 - 0.73 & bean.\textit{N} & 0.59 - 0.7 - 0.94 / 0.79 - 0.68 \\ 

cap.\textit{N} & 0.89 - 0.91 - 1.0 / 0.91 - 0.99 & teddy.\textit{N} & 0.52 - 0.45 - 0.91 / 0.48 - 0.64 & scale.\textit{N} & 0.53 - 0.59 - 0.95 / 0.59 - 0.82 \\ 

bra.\textit{N} & 0.52 - 0.5 - 0.9 / 0.48 - 0.45 & lump.\textit{N} & 0.52 - 0.87 - 0.97 / 0.91 - 0.87 & shuttle.\textit{N} & 0.85 - 0.85 - 0.84 / 0.85 - 0.7 \\ 

park.\textit{V} & 0.53 - 0.54 - 0.86 / 0.54 - 0.58 & comfort.\textit{N} & 0.78 - 0.78 - 0.78 / 0.78 - 0.77 & fuse.\textit{N} & 0.58 - 0.72 - 0.96 / 0.83 - 0.78 \\ 

drunk.\textit{JJ} & 0.56 - 0.69 - 0.77 / 0.68 - 0.76 & barn.\textit{N} & 0.75 - 0.75 - 0.78 / 0.78 - 0.79 & radioactive.\textit{JJ} & 0.54 - 0.54 - 0.98 / 0.69 - 0.63 \\ 

vegetable.\textit{N} & 0.59 - 0.61 - 0.74 / 0.61 - 0.71 & peanut.\textit{N} & 0.51 - 0.57 - 0.8 / 0.6 - 0.59 & pretend.\textit{V} & 0.56 - 0.54 - 0.81 / 0.57 - 0.69 \\ 

opening.\textit{N} & 0.55 - 0.6 - 0.85 / 0.59 - 0.7 & cabbage.\textit{N} & 0.63 - 0.63 - 0.7 / 0.65 - 0.74 & custom.\textit{N} & 0.52 - 0.65 - 0.99 / 0.81 - 0.98 \\ 

rule.\textit{V} & 0.68 - 0.7 - 0.87 / 0.72 - 0.88 & sandwich.\textit{N} & 0.62 - 0.62 - 0.61 / 0.62 - 0.6 & log.\textit{V} & 0.81 - 0.81 - 0.82 / 0.81 - 0.76 \\ 

rifle.\textit{N} & 0.7 - 0.71 - 0.96 / 0.78 - 0.85 & link.\textit{N} & 0.5 - 0.78 - 0.96 / 0.83 - 0.94 & riot.\textit{N} & 0.57 - 0.57 - 0.85 / 0.57 - 0.56 \\ 

herd.\textit{N} & 0.51 - 0.51 - 0.93 / 0.67 - 0.62 & supply.\textit{N} & 0.59 - 0.59 - 0.85 / 0.61 - 0.65 & sweater.\textit{N} & 0.56 - 0.54 - 0.56 / 0.56 - 0.47 \\ 

language.\textit{N} & 0.59 - 0.65 - 0.89 / 0.75 - 0.81 & privacy.\textit{N} & 0.56 - 0.58 - 0.9 / 0.6 - 0.66 & intrusion.\textit{N} & 0.65 - 0.61 - 0.64 / 0.65 - 0.65 \\ 

parking.\textit{N} & 0.68 - 0.65 - 0.67 / 0.68 - 0.65 & alien.\textit{JJ} & 0.64 - 0.64 - 0.96 / 0.64 - 0.6 & fighter.\textit{N} & 0.56 - 0.57 - 0.85 / 0.62 - 0.68 \\ 

approach.\textit{N} & 0.63 - 0.64 - 0.9 / 0.62 - 0.62 & pit.\textit{N} & 0.63 - 0.63 - 0.64 / 0.63 - 0.63 & cover.\textit{N} & 0.5 - 0.63 - 0.92 / 0.68 - 0.87 \\ 

bracelet.\textit{N} & 0.54 - 0.53 - 0.88 / 0.55 - 0.65 & gossip.\textit{N} & 0.55 - 0.55 - 0.81 / 0.57 - 0.65 & transfer.\textit{V} & 0.63 - 0.64 - 0.63 / 0.63 - 0.64 \\ 

horn.\textit{N} & 0.57 - 0.49 - 0.89 / 0.64 - 0.74 & dick.\textit{N} & 0.56 - 0.56 - 0.54 / 0.56 - 0.56 & reflection.\textit{N} & 0.6 - 0.67 - 0.98 / 0.79 - 0.86 \\ 

razor.\textit{N} & 0.69 - 0.71 - 0.89 / 0.76 - 0.68 & cleaner.\textit{N} & 0.56 - 0.98 - 1.0 / 0.98 - 0.98 & speaker.\textit{N} & 0.72 - 0.83 - 0.93 / 0.84 - 0.95 \\ 

computer.\textit{N} & 0.65 - 0.65 - 0.65 / 0.65 - 0.61 & survivor.\textit{N} & 0.55 - 0.56 - 0.81 / 0.59 - 0.47 & condolence.\textit{N} & 0.57 - 0.59 - 0.56 / 0.54 - 0.54 \\ 

flock.\textit{N} & 0.61 - 0.7 - 0.95 / 0.8 - 0.91 & dutch.\textit{JJ} & 0.69 - 0.81 - 0.98 / 0.83 - 0.86 & ounce.\textit{N} & 0.57 - 0.57 - 0.79 / 0.62 - 0.52 \\ 

cliff.\textit{N} & 0.69 - 0.31 - 0.7 / 0.69 - 0.74 & bite.\textit{N} & 0.57 - 0.66 - 0.85 / 0.64 - 0.77 & spread.\textit{V} & 0.37 - 0.44 - 0.95 / 0.52 - 0.51 \\ 

prayer.\textit{N} & 0.62 - 0.7 - 0.75 / 0.69 - 0.63 & match.\textit{N} & 0.52 - 0.52 - 0.52 / 0.52 - 0.56 & twenty.\textit{JJ} & 0.4 - 0.46 - 0.84 / 0.46 - 0.79 \\ 

promotion.\textit{N} & 0.56 - 0.57 - 0.89 / 0.6 - 0.68 & retire.\textit{V} & 0.52 - 0.52 - 0.88 / 0.64 - 0.69 & tribute.\textit{N} & 0.62 - 0.71 - 0.79 / 0.71 - 0.68 \\ 

vote.\textit{N} & 0.53 - 0.76 - 0.87 / 0.83 - 0.92 & honesty.\textit{N} & 0.66 - 0.66 - 0.66 / 0.66 - 0.59 & jar.\textit{N} & 0.59 - 0.59 - 0.6 / 0.59 - 0.62 \\ 

record.\textit{N} & 0.34 - 0.41 - 0.92 / 0.59 - 0.79 & twenty.\textit{N} & 0.74 - 0.76 - 0.77 / 0.74 - 0.93 & advance.\textit{N} & 0.36 - 0.41 - 0.87 / 0.5 - 0.7 \\ 

hunch.\textit{N} & 0.65 - 0.65 - 0.67 / 0.65 - 0.6 & praise.\textit{N} & 0.71 - 0.68 - 0.97 / 0.76 - 0.83 & jean.\textit{N} & 0.58 - 0.58 - 0.8 / 0.62 - 0.54 \\ 

skull.\textit{N} & 0.81 - 0.83 - 0.98 / 0.82 - 0.88 & wall.\textit{N} & 0.66 - 0.69 - 0.86 / 0.69 - 0.75 & restore.\textit{V} & 0.67 - 0.63 - 0.93 / 0.7 - 0.73 \\ 

essentially.\textit{RB} & 0.74 - 0.74 - 0.75 / 0.74 - 0.67 & mud.\textit{N} & 0.61 - 0.61 - 0.6 / 0.61 - 0.5 & alien.\textit{N} & 0.55 - 0.56 - 0.87 / 0.52 - 0.59 \\ 

requirement.\textit{N} & 0.7 - 0.7 - 0.72 / 0.7 - 0.62 & driver.\textit{N} & 0.63 - 0.68 - 0.68 / 0.68 - 0.69 & chin.\textit{N} & 0.64 - 0.64 - 0.93 / 0.56 - 0.59 \\ 

pneumonia.\textit{N} & 0.64 - 0.36 - 0.65 / 0.64 - 0.61 & relevant.\textit{JJ} & 0.66 - 0.66 - 0.98 / 0.68 - 0.61 & wave.\textit{N} & 0.66 - 0.73 - 0.89 / 0.78 - 0.89 \\ 

greed.\textit{N} & 0.57 - 0.57 - 0.98 / 0.51 - 0.49 & pill.\textit{N} & 0.46 - 0.49 - 0.72 / 0.52 - 0.59 & unfortunately.\textit{RB} & 0.36 - 0.36 - 0.59 / 0.34 - 0.42 \\ 

dagger.\textit{N} & 0.67 - 0.67 - 0.95 / 0.7 - 0.68 & riddle.\textit{N} & 0.71 - 0.29 - 0.71 / 0.71 - 0.7 & encourage.\textit{V} & 0.42 - 0.43 - 0.98 / 0.51 - 0.51 \\ 

editor.\textit{N} & 0.54 - 0.63 - 0.9 / 0.69 - 0.86 & rude.\textit{JJ} & 0.76 - 0.76 - 0.75 / 0.76 - 0.7 & belly.\textit{N} & 0.41 - 0.42 - 0.89 / 0.52 - 0.5 \\ 

maid.\textit{N} & 0.36 - 0.54 - 0.67 / 0.56 - 0.56 & calf.\textit{N} & 0.64 - 0.66 - 0.96 / 0.67 - 0.6 \\
temper.\textit{N} & 0.29 - 0.5 - 0.55 / 0.48 - 0.49 & ticket.\textit{N} & 0.58 - 0.64 - 0.77 / 0.67 - 0.66 \\
\midrule
 \multicolumn{3}{r|}{\textbf{Overall Average:}} & 59.43 - 62.40 - 66.87 - 70.72 \\
 \bottomrule
\end{tabular}
    }
    \caption{Lexical selection model test accuracies for all 157 English-Spanish words.}
   \label{tab:results}
     \end{center}
\end{table*}

\begin{table*}[t]
    \begin{center}
    \resizebox{\textwidth}{!}{
    \begin{tabular}{l|c|l|c|l|c}
Focus word & FreqBaseline - DTree - LinearSVM Train/Test - BERT & Focus word & FreqBaseline - DTree - LinearSVM Train/Test- BERT &Focus word & FreqBaseline - DTree - LinearSVM - BERT \\
\midrule
free.\textit{JJ} & 0.7 - 0.73 - 0.73 / 0.73 - 0.68 & peaceful.\textit{JJ} & 0.68 - 0.7 - 0.82 / 0.72 - 0.76 & fighter.\textit{N} & 0.29 - 0.31 - 0.78 / 0.31 - 0.47 \\ 

roof.\textit{N} & 0.37 - 0.39 - 0.63 / 0.39 - 0.46 & sharp.\textit{JJ} & 0.53 - 0.62 - 0.71 / 0.63 - 0.65 & crew.\textit{N} & 0.83 - 0.88 - 0.91 / 0.85 - 0.94 \\ 

sword.\textit{N} & 0.76 - 0.77 - 0.76 / 0.76 - 0.7 & tie.\textit{V} & 0.34 - 0.5 - 0.82 / 0.63 - 0.89 & convinced.\textit{JJ} & 0.66 - 0.68 - 0.73 / 0.66 - 0.65 \\ 

storm.\textit{N} & 0.85 - 0.84 - 0.87 / 0.84 - 0.83 & puzzle.\textit{N} & 0.61 - 0.69 - 0.81 / 0.71 - 0.7 & point.\textit{V} & 0.38 - 0.42 - 0.55 / 0.43 - 0.61 \\ 

break.\textit{V} & 0.16 - 0.42 - 0.73 / 0.55 - 0.75 & fan.\textit{N} & 0.61 - 0.65 - 0.87 / 0.69 - 0.8 & broke.\textit{JJ} & 0.37 - 0.4 - 0.72 / 0.35 - 0.4 \\ 

bitch.\textit{N} & 0.32 - 0.45 - 0.46 / 0.45 - 0.42 & shake.\textit{V} & 0.32 - 0.66 - 0.83 / 0.68 - 0.85 & sail.\textit{V} & 0.55 - 0.55 - 0.8 / 0.52 - 0.62 \\ 

set.\textit{V} & 0.3 - 0.48 - 0.59 / 0.51 - 0.71 & capital.\textit{N} & 0.71 - 0.78 - 0.9 / 0.77 - 0.98 & civil.\textit{JJ} & 0.53 - 0.84 - 0.96 / 0.86 - 0.91 \\ 

wheel.\textit{N} & 0.38 - 0.59 - 0.84 / 0.68 - 0.74 & sixth.\textit{JJ} & 0.78 - 0.78 - 0.78 / 0.78 - 0.71 & beef.\textit{N} & 0.37 - 0.37 - 0.77 / 0.39 - 0.43 \\ 

bone.\textit{N} & 0.33 - 0.46 - 0.85 / 0.62 - 0.82 & illusion.\textit{N} & 0.67 - 0.68 - 0.78 / 0.68 - 0.68 & deadline.\textit{N} & 0.8 - 0.8 - 0.8 / 0.8 - 0.76 \\ 

tunnel.\textit{N} & 0.67 - 0.67 - 0.68 / 0.67 - 0.61 & wet.\textit{JJ} & 0.41 - 0.68 - 0.77 / 0.68 - 0.71 & makeup.\textit{N} & 0.61 - 0.6 - 0.7 / 0.62 - 0.49 \\ 

cool.\textit{JJ} & 0.48 - 0.49 - 0.51 / 0.48 - 0.5 & costume.\textit{N} & 0.65 - 0.65 - 0.65 / 0.65 - 0.57 & text.\textit{N} & 0.84 - 0.84 - 0.92 / 0.84 - 0.86 \\ 

bedroom.\textit{N} & 0.63 - 0.63 - 0.67 / 0.64 - 0.6 & plague.\textit{N} & 0.69 - 0.68 - 0.84 / 0.69 - 0.67 & feed.\textit{V} & 0.32 - 0.46 - 0.88 / 0.48 - 0.76 \\ 

mountain.\textit{N} & 0.94 - 0.95 - 0.94 / 0.94 - 0.95 & vault.\textit{N} & 0.48 - 0.48 - 0.81 / 0.48 - 0.53 & bubble.\textit{N} & 0.54 - 0.57 - 0.86 / 0.55 - 0.56 \\ 

cake.\textit{N} & 0.93 - 0.99 - 0.98 / 0.99 - 0.99 & deadly.\textit{JJ} & 0.72 - 0.78 - 0.75 / 0.73 - 0.73 & drum.\textit{N} & 0.52 - 0.68 - 0.83 / 0.75 - 0.76 \\ 

coat.\textit{N} & 0.87 - 0.88 - 0.88 / 0.88 - 0.87 & collect.\textit{V} & 0.53 - 0.61 - 0.86 / 0.65 - 0.83 & drill.\textit{N} & 0.52 - 0.59 - 0.91 / 0.74 - 0.83 \\ 

bunch.\textit{N} & 0.78 - 0.79 - 0.84 / 0.79 - 0.73 & cliff.\textit{N} & 0.58 - 0.59 - 0.85 / 0.59 - 0.63 & musical.\textit{JJ} & 0.55 - 0.7 - 0.93 / 0.76 - 0.91 \\ 

turn.\textit{V} & 0.12 - 0.19 - 0.66 / 0.34 - 0.83 & scale.\textit{N} & 0.73 - 0.75 - 0.9 / 0.77 - 0.82 & burn.\textit{N} & 0.68 - 0.73 - 0.88 / 0.74 - 0.83 \\ 

effect.\textit{N} & 0.38 - 0.8 - 0.77 / 0.8 - 0.82 & horn.\textit{N} & 0.62 - 0.71 - 0.82 / 0.73 - 0.79 & lamb.\textit{N} & 0.77 - 0.76 - 0.86 / 0.77 - 0.77 \\ 

farm.\textit{N} & 0.95 - 0.95 - 0.95 / 0.95 - 0.94 & stick.\textit{N} & 0.46 - 0.52 - 0.52 / 0.51 - 0.47 & frame.\textit{N} & 0.37 - 0.35 - 0.97 / 0.57 - 0.57 \\ 

tie.\textit{N} & 0.67 - 0.76 - 0.89 / 0.81 - 0.93 & porn.\textit{N} & 0.79 - 0.79 - 0.79 / 0.79 - 0.77 & column.\textit{N} & 0.77 - 0.78 - 0.9 / 0.8 - 0.88 \\ 

tour.\textit{N} & 0.48 - 0.57 - 0.81 / 0.65 - 0.66 & range.\textit{N} & 0.58 - 0.66 - 0.86 / 0.67 - 0.69 & brilliant.\textit{JJ} & 0.43 - 0.43 - 0.56 / 0.44 - 0.51 \\ 

band.\textit{N} & 0.79 - 0.81 - 0.85 / 0.81 - 0.79 & host.\textit{N} & 0.56 - 0.64 - 0.88 / 0.74 - 0.87 & explode.\textit{V} & 0.51 - 0.59 - 0.85 / 0.58 - 0.9 \\ 

self.\textit{N} & 0.33 - 0.38 - 0.87 / 0.46 - 0.85 & grow.\textit{V} & 0.46 - 0.62 - 0.92 / 0.69 - 0.92 & fort.\textit{N} & 0.61 - 0.62 - 0.7 / 0.62 - 0.55 \\ 

bill.\textit{N} & 0.57 - 0.64 - 0.91 / 0.74 - 0.88 & expose.\textit{V} & 0.62 - 0.62 - 0.62 / 0.62 - 0.68 & impact.\textit{N} & 0.47 - 0.56 - 0.92 / 0.63 - 0.8 \\ 

cookie.\textit{N} & 0.79 - 0.77 - 0.8 / 0.79 - 0.73 & upset.\textit{JJ} & 0.47 - 0.53 - 0.54 / 0.53 - 0.46 & scarf.\textit{N} & 0.45 - 0.45 - 0.63 / 0.45 - 0.45 \\ 

dozen.\textit{N} & 0.58 - 0.66 - 0.83 / 0.71 - 0.81 & lightning.\textit{N} & 0.51 - 0.66 - 0.8 / 0.68 - 0.65 & fail.\textit{V} & 0.71 - 0.71 - 0.74 / 0.7 - 0.99 \\ 

fruit.\textit{N} & 0.79 - 0.86 - 0.95 / 0.88 - 0.93 & laptop.\textit{N} & 0.79 - 0.79 - 0.82 / 0.79 - 0.77 & skill.\textit{N} & 0.54 - 0.57 - 0.82 / 0.6 - 0.71 \\ 

pen.\textit{N} & 0.78 - 0.8 - 0.79 / 0.79 - 0.79 & response.\textit{N} & 0.57 - 0.59 - 0.61 / 0.58 - 0.64 & mail.\textit{N} & 0.9 - 0.95 - 0.97 / 0.96 - 0.95 \\ 

trigger.\textit{N} & 0.87 - 0.88 - 0.97 / 0.88 - 0.93 & obvious.\textit{JJ} & 0.52 - 0.54 - 0.6 / 0.55 - 0.58 & issue.\textit{V} & 0.5 - 0.54 - 0.86 / 0.58 - 0.79 \\ 

ring.\textit{N} & 0.42 - 0.63 - 0.79 / 0.72 - 0.81 & distant.\textit{JJ} & 0.67 - 0.77 - 0.96 / 0.89 - 0.92 & involve.\textit{V} & 0.28 - 0.31 - 0.69 / 0.32 - 0.38 \\ 

drag.\textit{V} & 0.27 - 0.33 - 0.65 / 0.43 - 0.64 & niece.\textit{N} & 0.76 - 0.76 - 0.76 / 0.76 - 0.72 & label.\textit{N} & 0.55 - 0.66 - 0.8 / 0.68 - 0.79 \\ 

old.\textit{JJ} & 0.4 - 0.66 - 0.86 / 0.73 - 0.91 & string.\textit{N} & 0.38 - 0.66 - 0.85 / 0.7 - 0.77 & psychic.\textit{JJ} & 0.61 - 0.74 - 0.93 / 0.79 - 0.84 \\ 

bug.\textit{N} & 0.35 - 0.42 - 0.54 / 0.46 - 0.52 & promote.\textit{V} & 0.59 - 0.62 - 0.92 / 0.75 - 0.83 & stamp.\textit{N} & 0.61 - 0.7 - 0.91 / 0.72 - 0.83 \\ 

campaign.\textit{N} & 0.53 - 0.53 - 0.83 / 0.53 - 0.58 & straight.\textit{JJ} & 0.43 - 0.53 - 0.86 / 0.61 - 0.76 & dump.\textit{N} & 0.42 - 0.62 - 0.73 / 0.64 - 0.7 \\ 

match.\textit{N} & 0.47 - 0.54 - 0.79 / 0.58 - 0.79 & worthy.\textit{JJ} & 0.75 - 0.75 - 0.75 / 0.75 - 0.81 & shell.\textit{N} & 0.36 - 0.53 - 0.9 / 0.56 - 0.7 \\ 

beat.\textit{V} & 0.22 - 0.29 - 0.53 / 0.34 - 0.54 & rocket.\textit{N} & 0.74 - 0.74 - 0.76 / 0.74 - 0.74 & disease.\textit{N} & 0.61 - 0.63 - 0.89 / 0.65 - 0.75 \\ 

bottom.\textit{N} & 0.69 - 0.78 - 0.85 / 0.78 - 0.8 & tub.\textit{N} & 0.7 - 0.89 - 0.88 / 0.89 - 0.87 & rule.\textit{V} & 0.33 - 0.4 - 0.9 / 0.59 - 0.76 \\ 

solve.\textit{V} & 0.41 - 0.51 - 0.67 / 0.53 - 0.66 & heir.\textit{N} & 0.66 - 0.68 - 0.87 / 0.68 - 0.67 & appeal.\textit{N} & 0.76 - 0.78 - 0.85 / 0.8 - 0.84 \\ 

sell.\textit{V} & 0.44 - 0.49 - 0.75 / 0.57 - 0.75 & circumstance.\textit{N} & 0.94 - 0.94 - 0.94 / 0.94 - 0.94 & can.\textit{N} & 0.55 - 0.68 - 0.87 / 0.7 - 0.71 \\ 

butter.\textit{N} & 0.59 - 0.82 - 0.86 / 0.82 - 0.93 & trunk.\textit{N} & 0.4 - 0.52 - 0.7 / 0.57 - 0.66 & glorious.\textit{JJ} & 0.74 - 0.74 - 0.75 / 0.74 - 0.7 \\ 

culture.\textit{N} & 0.48 - 0.53 - 0.89 / 0.58 - 0.64 & engagement.\textit{N} & 0.64 - 0.84 - 0.91 / 0.84 - 0.85 & focused.\textit{JJ} & 0.75 - 0.78 - 0.75 / 0.75 - 0.73 \\ 

season.\textit{N} & 0.62 - 0.67 - 0.81 / 0.68 - 0.66 & remain.\textit{N} & 0.37 - 0.38 - 0.91 / 0.35 - 0.38 & delicious.\textit{JJ} & 0.54 - 0.54 - 0.63 / 0.54 - 0.58 \\ 

tank.\textit{N} & 0.5 - 0.66 - 0.8 / 0.68 - 0.63 & crash.\textit{N} & 0.48 - 0.61 - 0.72 / 0.6 - 0.64 & addict.\textit{N} & 0.52 - 0.52 - 0.52 / 0.52 - 0.54 \\ 

wash.\textit{V} & 0.5 - 0.59 - 0.77 / 0.62 - 0.85 & cellar.\textit{N} & 0.84 - 0.84 - 0.84 / 0.84 - 0.84 & player.\textit{N} & 0.39 - 0.97 - 0.99 / 0.98 - 0.98 \\ 

ball.\textit{N} & 0.51 - 0.54 - 0.58 / 0.54 - 0.63 & opening.\textit{N} & 0.37 - 0.62 - 0.82 / 0.64 - 0.76 & lethal.\textit{JJ} & 0.75 - 0.79 - 0.85 / 0.79 - 0.78 \\ 

painful.\textit{JJ} & 0.48 - 0.49 - 0.5 / 0.49 - 0.46 & seal.\textit{N} & 0.6 - 0.66 - 0.95 / 0.77 - 0.9 & server.\textit{N} & 0.67 - 0.67 - 0.83 / 0.68 - 0.62 \\ 

general.\textit{JJ} & 0.68 - 0.69 - 0.95 / 0.78 - 0.99 & leak.\textit{V} & 0.56 - 0.58 - 0.85 / 0.6 - 0.76 & welcome.\textit{JJ} & 0.46 - 0.48 - 0.81 / 0.6 - 0.77 \\ 

evil.\textit{JJ} & 0.5 - 0.5 - 0.66 / 0.52 - 0.48 & harmless.\textit{JJ} & 0.51 - 0.56 - 0.56 / 0.55 - 0.49 & penny.\textit{N} & 0.93 - 0.92 - 0.93 / 0.92 - 0.98 \\ 

degree.\textit{N} & 0.5 - 0.62 - 0.93 / 0.86 - 0.93 & demand.\textit{N} & 0.66 - 0.76 - 0.95 / 0.8 - 0.94 & immune.\textit{JJ} & 0.57 - 0.93 - 0.94 / 0.93 - 0.97 \\ 

captain.\textit{N} & 0.52 - 0.54 - 0.52 / 0.52 - 0.67 & hip.\textit{N} & 0.65 - 0.66 - 0.86 / 0.68 - 0.85 & vet.\textit{N} & 0.73 - 0.8 - 0.94 / 0.82 - 0.91 \\ 

serial.\textit{JJ} & 0.63 - 0.84 - 0.85 / 0.85 - 0.82 & pride.\textit{N} & 0.61 - 0.64 - 0.8 / 0.65 - 0.85 & define.\textit{V} & 0.46 - 0.49 - 0.87 / 0.51 - 0.59 \\ 

infection.\textit{N} & 0.7 - 0.7 - 0.7 / 0.7 - 0.66 & file.\textit{V} & 0.4 - 0.43 - 0.57 / 0.46 - 0.58 & desperate.\textit{JJ} & 0.73 - 0.72 - 0.74 / 0.73 - 0.78 \\ 

fight.\textit{N} & 0.48 - 0.52 - 0.76 / 0.51 - 0.62 & burn.\textit{V} & 0.5 - 0.58 - 0.88 / 0.6 - 0.78 & move.\textit{V} & 0.31 - 0.47 - 0.9 / 0.62 - 0.89 \\ 

gut.\textit{N} & 0.56 - 0.74 - 0.87 / 0.79 - 0.84 & charm.\textit{N} & 0.75 - 0.85 - 0.88 / 0.85 - 0.88 & acre.\textit{N} & 0.72 - 0.72 - 0.72 / 0.72 - 0.72 \\ 

spring.\textit{N} & 0.92 - 0.92 - 0.92 / 0.92 - 0.97 & partner.\textit{N} & 0.53 - 0.63 - 0.93 / 0.72 - 0.88 & star.\textit{N} & 0.28 - 0.67 - 0.91 / 0.77 - 0.87 \\ 

spread.\textit{V} & 0.38 - 0.41 - 0.66 / 0.44 - 0.65 & youth.\textit{N} & 0.39 - 0.45 - 0.85 / 0.47 - 0.6 & claim.\textit{N} & 0.47 - 0.49 - 0.89 / 0.58 - 0.69 \\ 

hot.\textit{JJ} & 0.5 - 0.71 - 0.88 / 0.75 - 0.83 & raise.\textit{V} & 0.25 - 0.41 - 0.87 / 0.55 - 0.77 & leadership.\textit{N} & 0.77 - 0.81 - 0.87 / 0.8 - 0.81 \\ 

cup.\textit{N} & 0.7 - 0.68 - 0.72 / 0.7 - 0.67 & depressed.\textit{JJ} & 0.83 - 0.83 - 0.82 / 0.83 - 0.81 & collar.\textit{N} & 0.69 - 0.69 - 0.85 / 0.7 - 0.68 \\ 

clown.\textit{N} & 0.95 - 0.95 - 0.95 / 0.95 - 0.95 & toast.\textit{N} & 0.73 - 0.73 - 0.74 / 0.73 - 0.6 & donate.\textit{V} & 0.55 - 0.55 - 0.87 / 0.55 - 0.68 \\ 

lieutenant.\textit{N} & 0.41 - 0.48 - 0.71 / 0.52 - 0.63 & burger.\textit{N} & 0.52 - 0.5 - 0.52 / 0.52 - 0.44 & suspend.\textit{V} & 0.53 - 0.52 - 0.78 / 0.55 - 0.7 \\ 

original.\textit{JJ} & 0.46 - 0.59 - 0.82 / 0.65 - 0.77 & bury.\textit{V} & 0.74 - 0.76 - 0.76 / 0.75 - 0.88 & shade.\textit{N} & 0.58 - 0.76 - 0.94 / 0.88 - 0.92 \\ 

grass.\textit{N} & 0.51 - 0.52 - 0.83 / 0.58 - 0.64 & fatal.\textit{JJ} & 0.52 - 0.57 - 0.89 / 0.55 - 0.52 & sketch.\textit{N} & 0.79 - 0.81 - 0.8 / 0.79 - 0.93 \\ 

radiation.\textit{N} & 0.63 - 0.66 - 0.88 / 0.66 - 0.64 & abuse.\textit{N} & 0.61 - 0.72 - 0.79 / 0.72 - 0.77 & hood.\textit{N} & 0.55 - 0.7 - 0.94 / 0.82 - 0.88 \\ 

sad.\textit{JJ} & 0.5 - 0.64 - 0.85 / 0.73 - 0.81 & soft.\textit{JJ} & 0.75 - 0.82 - 0.88 / 0.82 - 0.87 & build.\textit{V} & 0.36 - 0.4 - 0.94 / 0.45 - 0.76 \\ 

necklace.\textit{N} & 0.79 - 0.79 - 0.79 / 0.79 - 0.76 & invade.\textit{V} & 0.74 - 0.74 - 0.74 / 0.74 - 0.86 & tear.\textit{V} & 0.39 - 0.49 - 0.86 / 0.51 - 0.75 \\ 

grade.\textit{N} & 0.42 - 0.68 - 0.88 / 0.89 - 0.9 & crack.\textit{N} & 0.46 - 0.59 - 0.91 / 0.64 - 0.83 & determine.\textit{V} & 0.62 - 0.63 - 0.93 / 0.7 - 0.94 \\ 

beast.\textit{N} & 0.66 - 0.67 - 0.8 / 0.67 - 0.66 & maid.\textit{N} & 0.59 - 0.77 - 0.8 / 0.78 - 0.74 & mourn.\textit{V} & 0.56 - 0.55 - 0.84 / 0.6 - 0.62 \\ 

blade.\textit{N} & 0.77 - 0.84 - 0.83 / 0.84 - 0.83 & spare.\textit{V} & 0.32 - 0.46 - 0.81 / 0.52 - 0.63 & abandon.\textit{V} & 0.48 - 0.55 - 0.86 / 0.59 - 0.75 \\ 

rise.\textit{V} & 0.23 - 0.43 - 0.88 / 0.6 - 0.82 & inappropriate.\textit{JJ} & 0.65 - 0.65 - 0.81 / 0.66 - 0.6 & lottery.\textit{N} & 0.51 - 0.63 - 0.78 / 0.61 - 0.59 \\ 

autopsy.\textit{JJ} & 0.54 - 0.53 - 0.79 / 0.51 - 0.48 & trouble.\textit{N} & 0.32 - 0.49 - 0.81 / 0.59 - 0.86 & pole.\textit{N} & 0.44 - 0.58 - 0.91 / 0.67 - 0.65 \\ 

jacket.\textit{N} & 0.65 - 0.65 - 0.66 / 0.66 - 0.55 & daily.\textit{JJ} & 0.7 - 0.83 - 0.96 / 0.83 - 0.83 & exhausted.\textit{JJ} & 0.62 - 0.6 - 0.61 / 0.62 - 0.49 \\ 

oil.\textit{N} & 0.83 - 0.9 - 0.95 / 0.89 - 0.88 & recording.\textit{N} & 0.6 - 0.56 - 0.82 / 0.59 - 0.59 & rubber.\textit{N} & 0.42 - 0.42 - 0.88 / 0.44 - 0.51 \\ 

compliment.\textit{N} & 0.57 - 0.26 - 0.58 / 0.57 - 0.51 & sink.\textit{N} & 0.7 - 0.7 - 0.7 / 0.7 - 0.68 & nipple.\textit{N} & 0.76 - 0.76 - 0.76 / 0.76 - 0.73 \\ 

pulse.\textit{N} & 0.7 - 0.76 - 0.78 / 0.77 - 0.82 & custom.\textit{N} & 0.58 - 0.66 - 0.89 / 0.71 - 0.97 & sew.\textit{V} & 0.48 - 0.52 - 0.87 / 0.56 - 0.72 \\ 

nephew.\textit{N} & 0.76 - 0.76 - 0.77 / 0.76 - 0.72 & brandy.\textit{N} & 0.76 - 0.76 - 0.76 / 0.76 - 0.74 & mental.\textit{JJ} & 0.44 - 0.38 - 0.84 / 0.45 - 0.47 \\ 

step.\textit{N} & 0.35 - 0.61 - 0.74 / 0.68 - 0.73 & souvenir.\textit{N} & 0.48 - 0.49 - 0.57 / 0.5 - 0.46 & janitor.\textit{N} & 0.8 - 0.8 - 0.8 / 0.8 - 0.77 \\ 

suffer.\textit{V} & 0.72 - 0.81 - 0.89 / 0.8 - 0.92 & pepper.\textit{N} & 0.67 - 0.84 - 0.9 / 0.84 - 0.97 & efficient.\textit{JJ} & 0.7 - 0.7 - 0.7 / 0.7 - 0.66 \\ 

run.\textit{V} & 0.19 - 0.32 - 0.84 / 0.59 - 0.84 & distraction.\textit{N} & 0.53 - 0.64 - 0.75 / 0.62 - 0.73 & speaker.\textit{N} & 0.45 - 0.51 - 0.84 / 0.58 - 0.8 \\ 

fight.\textit{V} & 0.21 - 0.29 - 0.61 / 0.34 - 0.65 & remarkable.\textit{JJ} & 0.44 - 0.5 - 0.78 / 0.52 - 0.52 & lawn.\textit{N} & 0.57 - 0.57 - 0.82 / 0.53 - 0.54 \\ 

store.\textit{N} & 0.24 - 0.66 - 0.81 / 0.69 - 0.7 & mob.\textit{N} & 0.58 - 0.62 - 0.91 / 0.69 - 0.86 & therapist.\textit{N} & 0.69 - 0.69 - 0.69 / 0.69 - 0.65 \\ 

fire.\textit{V} & 0.76 - 0.81 - 0.91 / 0.83 - 0.95 & casualty.\textit{N} & 0.78 - 0.78 - 0.78 / 0.78 - 0.7 & administration.\textit{N} & 0.52 - 0.52 - 0.9 / 0.56 - 0.67 \\ 

bright.\textit{JJ} & 0.71 - 0.86 - 0.89 / 0.86 - 0.87 & increase.\textit{V} & 0.63 - 0.66 - 0.96 / 0.68 - 0.9 & reckless.\textit{JJ} & 0.63 - 0.6 - 0.77 / 0.62 - 0.56 \\ 

inch.\textit{N} & 0.5 - 0.56 - 0.69 / 0.54 - 0.5 & melt.\textit{V} & 0.63 - 0.64 - 0.85 / 0.65 - 0.83 & ham.\textit{N} & 0.73 - 0.12 - 0.73 / 0.73 - 0.69 \\ 

barn.\textit{N} & 0.79 - 0.79 - 0.79 / 0.79 - 0.75 & thick.\textit{JJ} & 0.47 - 0.55 - 0.88 / 0.57 - 0.63 & settle.\textit{V} & 0.39 - 0.51 - 0.83 / 0.64 - 0.91 \\ 

gas.\textit{N} & 0.45 - 0.81 - 0.89 / 0.85 - 0.9 & mole.\textit{N} & 0.34 - 0.36 - 0.84 / 0.42 - 0.54 & headline.\textit{N} & 0.52 - 0.52 - 0.86 / 0.55 - 0.49 \\ 

cover.\textit{N} & 0.65 - 0.66 - 0.89 / 0.71 - 0.91 & football.\textit{N} & 0.39 - 0.41 - 0.75 / 0.49 - 0.52 & estate.\textit{N} & 0.43 - 0.71 - 0.82 / 0.71 - 0.76 \\ 

pot.\textit{N} & 0.38 - 0.49 - 0.68 / 0.6 - 0.56 & cattle.\textit{N} & 0.27 - 0.27 - 0.35 / 0.28 - 0.34 & smooth.\textit{JJ} & 0.38 - 0.45 - 0.88 / 0.45 - 0.6 \\ 

alley.\textit{N} & 0.44 - 0.44 - 0.81 / 0.49 - 0.51 & special.\textit{JJ} & 0.85 - 0.89 - 0.93 / 0.89 - 0.88 & worker.\textit{N} & 0.75 - 0.99 - 0.97 / 0.99 - 0.99 \\ 

liver.\textit{N} & 0.65 - 0.73 - 0.88 / 0.75 - 0.74 & high.\textit{JJ} & 0.6 - 0.66 - 0.69 / 0.66 - 0.63 & gallon.\textit{N} & 0.69 - 0.68 - 0.72 / 0.69 - 0.71 \\ 

escape.\textit{V} & 0.43 - 0.46 - 0.86 / 0.5 - 0.72 & clear.\textit{JJ} & 0.31 - 0.38 - 0.86 / 0.49 - 0.81 & scan.\textit{V} & 0.48 - 0.52 - 0.59 / 0.52 - 0.58 \\ 

beard.\textit{N} & 0.6 - 0.6 - 0.6 / 0.6 - 0.6 & paperwork.\textit{N} & 0.55 - 0.55 - 0.64 / 0.55 - 0.56 & lure.\textit{V} & 0.43 - 0.42 - 0.82 / 0.42 - 0.69 \\ 

moon.\textit{N} & 0.6 - 0.95 - 0.97 / 0.96 - 0.96 & dry.\textit{V} & 0.74 - 0.8 - 0.86 / 0.8 - 0.89 & sophisticated.\textit{JJ} & 0.35 - 0.35 - 0.66 / 0.37 - 0.45 \\ 

crown.\textit{N} & 0.81 - 0.85 - 0.96 / 0.86 - 0.89 & benefit.\textit{N} & 0.39 - 0.45 - 0.87 / 0.53 - 0.62 & offensive.\textit{JJ} & 0.72 - 0.82 - 0.92 / 0.81 - 0.89 \\ 

arrest.\textit{V} & 0.49 - 0.54 - 0.59 / 0.54 - 0.82 & scratch.\textit{N} & 0.56 - 0.59 - 0.57 / 0.56 - 0.56 & contribute.\textit{V} & 0.59 - 0.63 - 0.84 / 0.63 - 0.72 \\ 

male.\textit{JJ} & 0.6 - 0.61 - 0.86 / 0.63 - 0.69 & peanut.\textit{N} & 0.45 - 0.82 - 0.82 / 0.82 - 0.78 & management.\textit{N} & 0.55 - 0.6 - 0.89 / 0.61 - 0.7 \\ 

gum.\textit{N} & 0.5 - 0.5 - 0.77 / 0.52 - 0.69 & immunity.\textit{N} & 0.8 - 0.85 - 0.91 / 0.85 - 0.88 & straw.\textit{N} & 0.51 - 0.62 - 0.91 / 0.66 - 0.86 \\ 

approve.\textit{V} & 0.43 - 0.48 - 0.85 / 0.54 - 0.83 & cancel.\textit{V} & 0.83 - 0.83 - 0.83 / 0.83 - 0.85 & donkey.\textit{N} & 0.61 - 0.6 - 0.7 / 0.62 - 0.63 \\ 

candy.\textit{N} & 0.51 - 0.61 - 0.82 / 0.62 - 0.65 & wing.\textit{N} & 0.85 - 0.96 - 0.99 / 0.96 - 0.99 & delicate.\textit{JJ} & 0.7 - 0.7 - 0.7 / 0.7 - 0.68 \\ 

fifth.\textit{JJ} & 0.61 - 0.66 - 0.73 / 0.67 - 0.72 & camp.\textit{N} & 0.81 - 0.85 - 0.87 / 0.85 - 0.89 & brutal.\textit{JJ} & 0.39 - 0.42 - 0.43 / 0.4 - 0.28 \\ 

egg.\textit{N} & 0.63 - 0.78 - 0.86 / 0.81 - 0.89 & chip.\textit{N} & 0.87 - 1.0 - 0.99 / 1.0 - 1.0 & sunny.\textit{JJ} & 0.6 - 0.6 - 0.94 / 0.71 - 0.81 \\

 \bottomrule
\end{tabular}
    }
    \caption{Part-1: Lexical selection model test accuracies for a English-Greek words.}
   \label{tab:results-1}
     \end{center}
\end{table*}

\begin{table*}[t]
    \begin{center}
    \resizebox{\textwidth}{!}{
    \begin{tabular}{l|c|l|c|l|c}
Focus word & FreqBaseline - DTree - LinearSVM Train/Test - BERT & Focus word & FreqBaseline - DTree - LinearSVM Train/Test- BERT &Focus word & FreqBaseline - DTree - LinearSVM - BERT \\
\midrule
light.\textit{JJ} & 0.84 - 0.89 - 0.99 / 0.92 - 0.99 & camping.\textit{N} & 0.69 - 0.69 - 0.68 / 0.69 - 0.63 & suspension.\textit{N} & 0.42 - 0.48 - 0.96 / 0.58 - 0.73 \\ 

current.\textit{JJ} & 0.9 - 0.92 - 0.9 / 0.9 - 0.91 & cheat.\textit{V} & 0.34 - 0.33 - 0.64 / 0.3 - 0.67 & notorious.\textit{JJ} & 0.7 - 0.72 - 0.7 / 0.7 - 0.67 \\ 

humiliating.\textit{JJ} & 0.51 - 0.47 - 0.78 / 0.51 - 0.47 & act.\textit{V} & 0.51 - 0.53 - 0.87 / 0.51 - 0.65 & relative.\textit{JJ} & 0.59 - 0.64 - 0.98 / 0.76 - 0.95 \\ 

drone.\textit{N} & 0.38 - 0.39 - 0.89 / 0.41 - 0.52 & rose.\textit{N} & 0.47 - 0.51 - 0.9 / 0.57 - 0.76 & honor.\textit{V} & 0.82 - 0.88 - 0.99 / 0.93 - 0.93 \\ 

bald.\textit{JJ} & 0.69 - 0.69 - 0.69 / 0.69 - 0.64 & preacher.\textit{N} & 0.64 - 0.64 - 0.65 / 0.64 - 0.67 & duct.\textit{N} & 0.46 - 0.65 - 0.88 / 0.68 - 0.7 \\ 

wipe.\textit{V} & 0.38 - 0.4 - 0.85 / 0.46 - 0.55 & doorman.\textit{N} & 0.72 - 0.69 - 0.72 / 0.72 - 0.67 & scooter.\textit{N} & 0.78 - 0.78 - 0.79 / 0.78 - 0.72 \\ 

institution.\textit{N} & 0.54 - 0.67 - 0.89 / 0.68 - 0.76 & unite.\textit{V} & 0.53 - 0.53 - 0.91 / 0.52 - 0.7 & temporal.\textit{JJ} & 0.51 - 0.93 - 0.94 / 0.89 - 0.89 \\ 

retirement.\textit{N} & 0.54 - 0.58 - 0.87 / 0.63 - 0.57 & flock.\textit{N} & 0.51 - 0.54 - 0.88 / 0.56 - 0.59 & sis.\textit{N} & 0.55 - 0.55 - 0.85 / 0.58 - 0.42 \\ 

bunny.\textit{N} & 0.5 - 0.47 - 0.83 / 0.5 - 0.53 & remark.\textit{N} & 0.59 - 0.58 - 0.6 / 0.59 - 0.51 & terrace.\textit{N} & 0.63 - 0.63 - 0.63 / 0.63 - 0.7 \\ 

despair.\textit{N} & 0.53 - 0.53 - 0.95 / 0.55 - 0.56 & expel.\textit{V} & 0.5 - 0.62 - 0.68 / 0.62 - 0.74 & tenth.\textit{JJ} & 0.77 - 0.77 - 0.76 / 0.77 - 0.73 \\ 

cult.\textit{N} & 0.75 - 0.75 - 0.75 / 0.75 - 0.71 & dizzy.\textit{JJ} & 0.45 - 0.51 - 0.66 / 0.53 - 0.56 & banner.\textit{N} & 0.5 - 0.52 - 0.91 / 0.55 - 0.67 \\ 

trainer.\textit{N} & 0.44 - 0.54 - 0.83 / 0.51 - 0.57 & vulture.\textit{N} & 0.53 - 0.19 - 0.83 / 0.62 - 0.62 & mixture.\textit{N} & 0.53 - 0.54 - 0.95 / 0.5 - 0.41 \\ 

genius.\textit{N} & 0.66 - 0.66 - 0.81 / 0.66 - 0.67 & simulation.\textit{N} & 0.78 - 0.78 - 0.79 / 0.78 - 0.79 & swelling.\textit{N} & 0.73 - 0.71 - 0.79 / 0.7 - 0.73 \\ 

competition.\textit{N} & 0.87 - 0.86 - 0.87 / 0.87 - 0.87 & contempt.\textit{N} & 0.74 - 0.85 - 0.88 / 0.86 - 0.81 & clip.\textit{N} & 0.37 - 0.68 - 0.93 / 0.83 - 0.83 \\ 

thread.\textit{N} & 0.63 - 0.65 - 0.81 / 0.64 - 0.67 & dean.\textit{N} & 0.8 - 0.81 - 0.8 / 0.8 - 0.76 & mix.\textit{V} & 0.57 - 0.6 - 0.86 / 0.64 - 0.84 \\ 

coach.\textit{N} & 0.83 - 0.82 - 0.84 / 0.83 - 0.94 & beg.\textit{V} & 0.79 - 0.82 - 0.94 / 0.8 - 0.91 & eliminate.\textit{V} & 0.46 - 0.51 - 0.52 / 0.46 - 0.63 \\ 

trance.\textit{N} & 0.82 - 0.84 - 0.82 / 0.82 - 0.99 & grieve.\textit{V} & 0.67 - 0.67 - 0.69 / 0.67 - 0.63 & bouquet.\textit{N} & 0.63 - 0.63 - 0.73 / 0.64 - 0.57 \\ 

respectable.\textit{JJ} & 0.52 - 0.51 - 0.53 / 0.52 - 0.34 & pea.\textit{N} & 0.59 - 0.56 - 0.78 / 0.59 - 0.62 & dig.\textit{V} & 0.53 - 0.63 - 0.65 / 0.63 - 0.85 \\ 

notebook.\textit{N} & 0.78 - 0.78 - 0.78 / 0.78 - 0.78 & extension.\textit{N} & 0.38 - 0.51 - 0.94 / 0.56 - 0.68 & abs.\textit{N} & 0.55 - 0.58 - 0.93 / 0.64 - 0.97 \\ 

recommendation.\textit{N} & 0.54 - 0.73 - 0.83 / 0.74 - 0.7 & intercept.\textit{V} & 0.55 - 0.65 - 0.87 / 0.72 - 0.87 & desperation.\textit{N} & 0.59 - 0.59 - 0.9 / 0.52 - 0.48 \\ 

wire.\textit{N} & 0.61 - 0.71 - 0.93 / 0.73 - 0.77 & homemade.\textit{JJ} & 0.77 - 0.79 - 0.77 / 0.77 - 0.85 & slogan.\textit{N} & 0.77 - 0.77 - 0.77 / 0.77 - 0.79 \\ 

humiliation.\textit{N} & 0.6 - 0.62 - 0.69 / 0.63 - 0.56 & domestic.\textit{JJ} & 0.48 - 0.7 - 0.93 / 0.62 - 0.75 & raven.\textit{N} & 0.65 - 0.68 - 0.88 / 0.65 - 0.89 \\ 

dock.\textit{N} & 0.48 - 0.61 - 0.8 / 0.6 - 0.74 & spear.\textit{N} & 0.43 - 0.42 - 0.8 / 0.38 - 0.58 & waste.\textit{V} & 0.58 - 0.86 - 0.96 / 0.83 - 0.94 \\ 

recover.\textit{V} & 0.38 - 0.48 - 0.87 / 0.52 - 0.79 & printer.\textit{N} & 0.78 - 0.77 - 0.78 / 0.78 - 0.86 & honorable.\textit{JJ} & 0.75 - 0.79 - 0.97 / 0.77 - 0.75 \\ 

fortress.\textit{N} & 0.71 - 0.69 - 0.71 / 0.71 - 0.61 & carnival.\textit{N} & 0.72 - 0.75 - 0.72 / 0.72 - 0.64 & clarity.\textit{N} & 0.69 - 0.69 - 0.7 / 0.69 - 0.69 \\ 

furious.\textit{JJ} & 0.67 - 0.67 - 0.68 / 0.67 - 0.62 & intervene.\textit{V} & 0.73 - 0.7 - 0.73 / 0.73 - 0.68 & minority.\textit{N} & 0.66 - 0.66 - 0.91 / 0.67 - 0.72 \\ 

light.\textit{V} & 0.62 - 0.73 - 0.94 / 0.74 - 0.83 & concrete.\textit{N} & 0.63 - 0.65 - 0.82 / 0.67 - 0.7 & frustrating.\textit{JJ} & 0.73 - 0.73 - 0.76 / 0.73 - 0.61 \\ 

sign.\textit{V} & 0.82 - 0.82 - 0.83 / 0.82 - 0.98 & argument.\textit{N} & 0.58 - 0.58 - 0.57 / 0.58 - 0.55 & resident.\textit{N} & 0.7 - 0.73 - 0.69 / 0.7 - 0.85 \\ 

crop.\textit{N} & 0.63 - 0.69 - 0.86 / 0.69 - 0.63 & extensive.\textit{JJ} & 0.72 - 0.72 - 0.72 / 0.72 - 0.71 & spaghetti.\textit{N} & 0.56 - 0.56 - 0.55 / 0.56 - 0.48 \\ 

spill.\textit{V} & 0.56 - 0.65 - 0.85 / 0.7 - 0.81 & kiss.\textit{V} & 0.36 - 0.68 - 0.92 / 0.77 - 0.89 & relic.\textit{N} & 0.6 - 0.6 - 0.78 / 0.58 - 0.75 \\ 

congressman.\textit{N} & 0.59 - 0.62 - 0.74 / 0.61 - 0.52 & harvest.\textit{N} & 0.61 - 0.64 - 0.74 / 0.64 - 0.71 & shaft.\textit{N} & 0.5 - 0.65 - 0.83 / 0.62 - 0.75 \\ 

sale.\textit{N} & 0.53 - 0.71 - 0.92 / 0.76 - 0.99 & foreman.\textit{N} & 0.43 - 0.51 - 0.81 / 0.58 - 0.77 & breathe.\textit{RB} & 0.67 - 0.67 - 0.6 / 0.67 - 0.61 \\ 

advanced.\textit{JJ} & 0.67 - 0.73 - 0.84 / 0.73 - 0.78 & pier.\textit{N} & 0.82 - 0.82 - 0.82 / 0.82 - 0.78 & contraction.\textit{N} & 0.69 - 0.69 - 0.69 / 0.69 - 0.67 \\ 

publish.\textit{V} & 0.52 - 0.53 - 0.62 / 0.55 - 0.74 & ignorant.\textit{JJ} & 0.6 - 0.56 - 0.6 / 0.6 - 0.5 & outbreak.\textit{N} & 0.6 - 0.6 - 0.6 / 0.6 - 0.53 \\ 

popcorn.\textit{N} & 0.77 - 0.23 - 0.77 / 0.77 - 0.71 & fart.\textit{V} & 0.68 - 0.73 - 0.68 / 0.68 - 0.9 & record.\textit{V} & 0.83 - 0.85 - 0.85 / 0.83 - 0.97 \\ 

thunder.\textit{N} & 0.62 - 0.61 - 0.68 / 0.62 - 0.68 & sabotage.\textit{N} & 0.78 - 0.78 - 0.78 / 0.78 - 0.68 & ranger.\textit{N} & 0.55 - 0.7 - 0.86 / 0.73 - 0.78 \\ 

wreck.\textit{N} & 0.35 - 0.39 - 0.51 / 0.38 - 0.58 & stripe.\textit{N} & 0.54 - 0.63 - 0.89 / 0.65 - 0.59 & cathedral.\textit{N} & 0.55 - 0.52 - 0.86 / 0.57 - 0.57 \\ 

shine.\textit{V} & 0.33 - 0.65 - 0.88 / 0.71 - 0.84 & rooftop.\textit{N} & 0.66 - 0.66 - 0.66 / 0.66 - 0.66 & boredom.\textit{N} & 0.48 - 0.48 - 0.52 / 0.48 - 0.5 \\ 

bite.\textit{N} & 0.76 - 0.87 - 0.97 / 0.83 - 0.92 & destroyer.\textit{N} & 0.59 - 0.64 - 0.95 / 0.79 - 0.86 & manual.\textit{JJ} & 0.73 - 1.0 - 1.0 / 1.0 - 1.0 \\ 

contaminate.\textit{V} & 0.44 - 0.43 - 0.57 / 0.43 - 0.8 & whine.\textit{V} & 0.41 - 0.41 - 0.53 / 0.39 - 0.32 & interfere.\textit{V} & 0.76 - 0.78 - 0.75 / 0.76 - 0.85 \\ 

intern.\textit{N} & 0.56 - 0.56 - 0.57 / 0.56 - 0.55 & muffin.\textit{N} & 0.59 - 0.59 - 0.76 / 0.6 - 0.55 & quality.\textit{N} & 0.57 - 0.55 - 0.84 / 0.62 - 0.58 \\ 

willing.\textit{JJ} & 0.69 - 0.83 - 0.82 / 0.83 - 0.77 & delusion.\textit{N} & 0.59 - 0.59 - 0.59 / 0.59 - 0.55 & obstruction.\textit{N} & 0.65 - 0.56 - 0.66 / 0.65 - 0.56 \\ 

fountain.\textit{N} & 0.63 - 0.63 - 0.69 / 0.63 - 0.59 & tornado.\textit{N} & 0.76 - 0.76 - 0.76 / 0.76 - 0.77 & modification.\textit{N} & 0.65 - 0.67 - 0.69 / 0.65 - 0.72 \\ 

compete.\textit{V} & 0.56 - 0.58 - 0.86 / 0.62 - 0.71 & rock.\textit{N} & 0.81 - 0.96 - 1.0 / 0.96 - 1.0 & marrow.\textit{N} & 0.71 - 0.72 - 0.89 / 0.72 - 0.8 \\ 

mentally.\textit{RB} & 0.59 - 0.71 - 0.82 / 0.73 - 0.73 & courier.\textit{N} & 0.52 - 0.52 - 0.89 / 0.58 - 0.48 & loser.\textit{N} & 0.66 - 0.66 - 0.68 / 0.66 - 0.75 \\ 

swim.\textit{N} & 0.5 - 0.71 - 0.84 / 0.68 - 0.76 & rat.\textit{N} & 0.41 - 0.67 - 0.86 / 0.68 - 0.86 & branch.\textit{N} & 0.53 - 0.54 - 0.9 / 0.59 - 0.9 \\ 

birth.\textit{N} & 0.37 - 0.82 - 0.84 / 0.83 - 0.89 & gather.\textit{V} & 0.62 - 0.83 - 0.95 / 0.9 - 0.9 & bankruptcy.\textit{N} & 0.5 - 0.59 - 0.89 / 0.61 - 0.56 \\ 

sequence.\textit{N} & 0.72 - 0.77 - 0.85 / 0.78 - 0.74 & countryside.\textit{N} & 0.65 - 0.71 - 0.65 / 0.65 - 0.79 & profitable.\textit{JJ} & 0.54 - 0.48 - 0.91 / 0.58 - 0.5 \\ 

dirty.\textit{JJ} & 0.4 - 0.84 - 0.97 / 0.87 - 0.93 & train.\textit{N} & 0.49 - 0.74 - 0.94 / 0.86 - 0.94 & broker.\textit{N} & 0.62 - 0.67 - 0.74 / 0.67 - 0.65 \\ 

attempt.\textit{V} & 0.8 - 0.8 - 0.8 / 0.8 - 0.85 & guest.\textit{N} & 0.52 - 0.74 - 0.89 / 0.76 - 0.87 & official.\textit{N} & 0.69 - 0.7 - 0.96 / 0.69 - 0.8 \\ 

link.\textit{N} & 0.57 - 0.74 - 0.96 / 0.78 - 0.94 & tonic.\textit{N} & 0.78 - 0.92 - 0.83 / 0.9 - 0.99 & distract.\textit{V} & 0.58 - 0.58 - 0.84 / 0.65 - 0.87 \\ 

request.\textit{N} & 0.55 - 0.6 - 0.8 / 0.61 - 0.68 & yen.\textit{N} & 0.67 - 0.65 - 0.65 / 0.67 - 0.67 & remote.\textit{N} & 0.53 - 0.49 - 0.86 / 0.53 - 0.53 \\ 

cautious.\textit{JJ} & 0.79 - 0.79 - 0.78 / 0.79 - 0.73 & equal.\textit{V} & 0.56 - 0.61 - 0.6 / 0.57 - 0.62 & ignition.\textit{N} & 0.77 - 0.82 - 0.95 / 0.79 - 0.89 \\ 

accessory.\textit{N} & 0.61 - 0.71 - 0.96 / 0.7 - 0.88 & isolated.\textit{JJ} & 0.7 - 0.94 - 0.91 / 0.94 - 0.97 & weed.\textit{N} & 0.47 - 0.57 - 0.92 / 0.59 - 0.71 \\ 

ounce.\textit{N} & 0.58 - 0.53 - 0.8 / 0.53 - 0.58 & arrogance.\textit{N} & 0.78 - 0.78 - 0.78 / 0.78 - 0.73 & vigilante.\textit{N} & 0.7 - 0.7 - 0.7 / 0.7 - 0.68 \\ 

spinal.\textit{JJ} & 0.52 - 0.68 - 0.81 / 0.71 - 0.62 & native.\textit{N} & 0.51 - 0.39 - 0.79 / 0.39 - 0.54 & proportion.\textit{N} & 0.62 - 0.62 - 0.98 / 0.7 - 0.81 \\ 

editor.\textit{N} & 0.59 - 0.64 - 0.92 / 0.59 - 0.65 & redemption.\textit{N} & 0.81 - 0.19 - 0.8 / 0.81 - 0.74 & pedophile.\textit{N} & 0.63 - 0.63 - 0.72 / 0.63 - 0.58 \\ 

shocking.\textit{JJ} & 0.75 - 0.75 - 0.74 / 0.75 - 0.69 & rookie.\textit{N} & 0.36 - 0.36 - 0.65 / 0.35 - 0.27 & scenery.\textit{N} & 0.68 - 0.88 - 0.91 / 0.8 - 0.88 \\ 

miserable.\textit{JJ} & 0.74 - 0.75 - 0.75 / 0.74 - 0.73 & robber.\textit{N} & 0.38 - 0.64 - 0.76 / 0.67 - 0.67 & ballot.\textit{N} & 0.55 - 0.68 - 0.98 / 0.7 - 0.64 \\ 

scan.\textit{N} & 0.66 - 0.77 - 0.86 / 0.78 - 0.73 & decency.\textit{N} & 0.59 - 0.56 - 0.86 / 0.54 - 0.54 & nightfall.\textit{N} & 0.52 - 0.5 - 0.9 / 0.55 - 0.78 \\ 

rider.\textit{N} & 0.6 - 0.62 - 0.85 / 0.64 - 0.7 & mold.\textit{N} & 0.63 - 0.76 - 0.93 / 0.77 - 0.97 & lookout.\textit{N} & 0.53 - 0.53 - 0.83 / 0.56 - 0.78 \\ 

choose.\textit{V} & 0.59 - 0.68 - 0.84 / 0.7 - 0.87 & brothel.\textit{N} & 0.52 - 0.52 - 0.84 / 0.48 - 0.52 & violet.\textit{N} & 0.62 - 0.66 - 0.98 / 0.75 - 0.94 \\ 

push.\textit{V} & 0.54 - 0.58 - 0.61 / 0.58 - 0.7 & breathe.\textit{V} & 0.64 - 0.72 - 0.9 / 0.7 - 0.86 & caleb.\textit{JJ} & 0.74 - 0.74 - 0.74 / 0.74 - 0.65 \\ 

pajama.\textit{N} & 0.6 - 0.6 - 0.6 / 0.6 - 0.58 & deliberately.\textit{RB} & 0.69 - 0.68 - 0.7 / 0.69 - 0.64 & static.\textit{JJ} & 0.65 - 0.72 - 0.9 / 0.69 - 0.89 \\ 

thorough.\textit{JJ} & 0.39 - 0.43 - 0.6 / 0.4 - 0.34 & adjustment.\textit{N} & 0.63 - 0.65 - 0.81 / 0.73 - 0.72 & baptize.\textit{V} & 0.55 - 0.59 - 0.86 / 0.55 - 0.62 \\ 

stubborn.\textit{JJ} & 0.86 - 0.86 - 0.86 / 0.86 - 0.85 & component.\textit{N} & 0.51 - 0.61 - 0.82 / 0.54 - 0.87 & mark.\textit{V} & 0.52 - 0.65 - 0.92 / 0.72 - 0.83 \\ 

penetrate.\textit{V} & 0.57 - 0.57 - 0.93 / 0.65 - 0.69 & lust.\textit{N} & 0.58 - 0.58 - 0.81 / 0.6 - 0.57 & modest.\textit{JJ} & 0.58 - 0.51 - 0.83 / 0.55 - 0.58 \\ 

guard.\textit{N} & 0.23 - 0.69 - 0.95 / 0.77 - 0.93 & classy.\textit{JJ} & 0.38 - 0.38 - 0.47 / 0.38 - 0.27 & insight.\textit{N} & 0.42 - 0.42 - 0.81 / 0.46 - 0.5 \\ 

decorate.\textit{V} & 0.59 - 0.77 - 0.89 / 0.78 - 0.83 & unsolved.\textit{JJ} & 0.63 - 0.63 - 0.63 / 0.63 - 0.67 & chart.\textit{N} & 0.86 - 0.86 - 1.0 / 0.88 - 0.94 \\ 

cord.\textit{N} & 0.44 - 0.77 - 0.9 / 0.8 - 0.8 & bravery.\textit{N} & 0.66 - 0.7 - 0.95 / 0.68 - 0.59 & shrink.\textit{N} & 0.6 - 0.6 - 0.58 / 0.6 - 0.53 \\ 

eighth.\textit{JJ} & 0.76 - 0.76 - 0.76 / 0.76 - 0.7 & radius.\textit{N} & 0.29 - 0.46 - 0.88 / 0.66 - 0.8 & whorehouse.\textit{N} & 0.51 - 0.46 - 0.94 / 0.41 - 0.56 \\ 

poll.\textit{N} & 0.6 - 0.79 - 0.89 / 0.77 - 0.78 & register.\textit{V} & 0.29 - 0.38 - 0.93 / 0.47 - 0.73 & memorial.\textit{N} & 0.53 - 0.58 - 0.98 / 0.77 - 0.72 \\ 

pathetic.\textit{JJ} & 0.49 - 0.27 - 0.53 / 0.49 - 0.44 & paint.\textit{V} & 0.61 - 0.65 - 0.93 / 0.61 - 0.89 & mineral.\textit{N} & 0.52 - 0.85 - 0.9 / 0.85 - 0.89 \\ 

prey.\textit{N} & 0.55 - 0.61 - 0.91 / 0.64 - 0.6 & powerless.\textit{JJ} & 0.53 - 0.47 - 0.81 / 0.51 - 0.65 & tracker.\textit{N} & 0.45 - 0.5 - 0.67 / 0.52 - 0.69 \\ 

settlement.\textit{N} & 0.63 - 0.63 - 0.89 / 0.64 - 0.79 & exhaust.\textit{V} & 0.78 - 0.78 - 0.78 / 0.78 - 0.65 & rebuild.\textit{V} & 0.6 - 0.6 - 0.97 / 0.75 - 0.64 \\ 

bow.\textit{N} & 0.33 - 0.64 - 0.9 / 0.72 - 0.88 & bend.\textit{V} & 0.3 - 0.53 - 0.88 / 0.65 - 0.76 & consumption.\textit{N} & 0.81 - 0.83 - 0.97 / 0.83 - 0.87 \\ 

dorm.\textit{N} & 0.69 - 0.69 - 0.68 / 0.69 - 0.64 & porter.\textit{N} & 0.41 - 0.41 - 0.79 / 0.43 - 0.65 & viper.\textit{N} & 0.57 - 0.57 - 0.97 / 0.61 - 0.78 \\ 

serve.\textit{V} & 0.66 - 0.66 - 0.9 / 0.74 - 0.89 & guinea.\textit{N} & 0.52 - 1.0 - 0.98 / 1.0 - 1.0 & milligram.\textit{N} & 0.73 - 0.76 - 0.76 / 0.73 - 0.64 \\ 

contribution.\textit{N} & 0.7 - 0.7 - 0.7 / 0.7 - 0.67 & serpent.\textit{N} & 0.58 - 0.42 - 0.56 / 0.58 - 0.52 & hanging.\textit{N} & 0.72 - 0.76 - 0.96 / 0.85 - 0.65 \\ 

grocery.\textit{N} & 0.36 - 0.4 - 0.73 / 0.41 - 0.38 & purity.\textit{N} & 0.77 - 0.77 - 0.78 / 0.77 - 0.79 & expansion.\textit{N} & 0.83 - 0.88 - 0.95 / 0.94 - 0.88 \\ 

sunshine.\textit{N} & 0.48 - 0.57 - 0.8 / 0.63 - 0.65 & college.\textit{N} & 0.44 - 0.56 - 0.91 / 0.71 - 0.78 & reluctant.\textit{JJ} & 0.64 - 0.62 - 0.68 / 0.64 - 0.6 \\ 

sponsor.\textit{N} & 0.56 - 0.56 - 0.78 / 0.59 - 0.65 & guarantee.\textit{V} & 0.68 - 0.68 - 0.67 / 0.68 - 0.86 & declaration.\textit{N} & 0.56 - 0.82 - 0.94 / 0.82 - 0.71 \\ 

broken.\textit{JJ} & 0.43 - 0.86 - 0.94 / 0.86 - 0.9 & camp.\textit{V} & 0.68 - 0.71 - 0.86 / 0.72 - 0.84 & regional.\textit{JJ} & 0.6 - 0.64 - 0.89 / 0.6 - 0.55 \\ 

convoy.\textit{N} & 0.28 - 0.28 - 0.79 / 0.33 - 0.39 & promising.\textit{JJ} & 0.77 - 0.77 - 0.77 / 0.77 - 0.61 & sterile.\textit{JJ} & 0.6 - 0.6 - 0.92 / 0.69 - 0.9 \\ 

unbearable.\textit{JJ} & 0.44 - 0.46 - 0.66 / 0.49 - 0.39 & detector.\textit{N} & 0.54 - 0.8 - 0.87 / 0.83 - 0.79 & skinny.\textit{JJ} & 0.65 - 0.62 - 0.66 / 0.65 - 0.56 \\ 

vague.\textit{JJ} & 0.6 - 0.6 - 0.6 / 0.6 - 0.56 & elect.\textit{V} & 0.6 - 0.61 - 0.82 / 0.63 - 0.73 & amendment.\textit{N} & 0.57 - 0.57 - 0.8 / 0.57 - 0.61 \\ 

torch.\textit{N} & 0.44 - 0.44 - 0.44 / 0.44 - 0.44 & paramedic.\textit{N} & 0.41 - 0.42 - 0.75 / 0.44 - 0.58 & binocular.\textit{N} & 0.65 - 0.65 - 0.64 / 0.65 - 0.53 \\ 

boxer.\textit{N} & 0.76 - 0.76 - 0.77 / 0.76 - 0.74 & shiny.\textit{JJ} & 0.79 - 0.79 - 0.79 / 0.79 - 0.76 & crutch.\textit{N} & 0.61 - 0.59 - 0.7 / 0.61 - 0.64 \\ 

chin.\textit{N} & 0.61 - 0.23 - 0.61 / 0.61 - 0.59 & racial.\textit{JJ} & 0.56 - 0.69 - 0.85 / 0.69 - 0.56 & grill.\textit{N} & 0.47 - 0.47 - 0.64 / 0.47 - 0.49 \\ 

cube.\textit{N} & 0.61 - 0.95 - 0.97 / 0.95 - 0.95 & vagina.\textit{N} & 0.55 - 0.55 - 0.55 / 0.55 - 0.48 & cone.\textit{N} & 0.53 - 0.65 - 0.89 / 0.69 - 0.78 \\ 

lean.\textit{V} & 0.3 - 0.34 - 0.85 / 0.39 - 0.65 & poor.\textit{JJ} & 0.74 - 0.74 - 0.73 / 0.74 - 0.71 & militia.\textit{N} & 0.64 - 0.64 - 0.65 / 0.64 - 0.68 \\ 

carrier.\textit{N} & 0.59 - 0.65 - 0.94 / 0.73 - 0.95 & neural.\textit{JJ} & 0.54 - 0.58 - 0.93 / 0.68 - 0.65 & count.\textit{V} & 0.42 - 0.42 - 0.86 / 0.49 - 0.91 \\ 

release.\textit{V} & 0.56 - 0.68 - 0.91 / 0.74 - 0.93 & substitute.\textit{N} & 0.57 - 0.64 - 0.79 / 0.65 - 0.74 & voluntarily.\textit{RB} & 0.63 - 0.37 - 0.63 / 0.63 - 0.54 \\ 

dot.\textit{N} & 0.58 - 0.65 - 0.89 / 0.78 - 0.81 & bolt.\textit{N} & 0.29 - 0.56 - 0.69 / 0.57 - 0.54 & gallow.\textit{N} & 0.62 - 0.66 - 0.73 / 0.66 - 0.5 \\ 

pantie.\textit{N} & 0.46 - 0.46 - 0.66 / 0.46 - 0.44 & oral.\textit{JJ} & 0.74 - 0.88 - 0.91 / 0.75 - 0.96 & hangover.\textit{N} & 0.6 - 0.6 - 0.77 / 0.6 - 0.53 \\ 

roadblock.\textit{N} & 0.66 - 0.66 - 0.66 / 0.66 - 0.53 & forensic.\textit{JJ} & 0.51 - 0.58 - 0.94 / 0.65 - 0.5 & operating.\textit{N} & 0.74 - 1.0 - 1.0 / 1.0 - 1.0 \\ 

blackout.\textit{N} & 0.62 - 0.66 - 0.82 / 0.66 - 0.53 & stench.\textit{N} & 0.55 - 0.55 - 0.81 / 0.55 - 0.51 & pasta.\textit{N} & 0.56 - 0.56 - 0.9 / 0.6 - 0.55 \\ 

superstition.\textit{N} & 0.37 - 0.37 - 0.5 / 0.37 - 0.33 & carpenter.\textit{N} & 0.72 - 0.72 - 0.73 / 0.72 - 0.63 & overrate.\textit{V} & 0.53 - 0.53 - 0.82 / 0.53 - 0.61 \\ 

silver.\textit{N} & 0.85 - 0.91 - 0.98 / 0.92 - 0.97 & mri.\textit{N} & 0.78 - 0.78 - 0.77 / 0.78 - 0.78 & charge.\textit{V} & 0.6 - 0.88 - 0.97 / 0.9 - 0.93 \\ 

complex.\textit{JJ} & 0.53 - 0.54 - 0.96 / 0.54 - 0.66 & deport.\textit{V} & 0.63 - 0.63 - 0.68 / 0.65 - 0.66 & rift.\textit{N} & 0.68 - 0.78 - 0.78 / 0.73 - 0.84 \\

 \bottomrule
\end{tabular}
    }
    \caption{Part-2: Lexical selection model test accuracies for a English-Greek words.}
   \label{tab:results-2}
     \end{center}
\end{table*}

\begin{table*}[t]
    \begin{center}
    \resizebox{\textwidth}{!}{
    \begin{tabular}{l|c|l|c|l|c}
Focus word & FreqBaseline - DTree - LinearSVM Train/Test - BERT & Focus word & FreqBaseline - DTree - LinearSVM Train/Test- BERT &Focus word & FreqBaseline - DTree - LinearSVM - BERT \\
\midrule
descent.\textit{N} & 0.64 - 0.64 - 0.94 / 0.67 - 0.86 & 
suit.\textit{N} & 0.63 - 0.63 - 0.95 / 0.7 - 1.0 & 
shrine.\textit{N} & 0.51 - 0.51 - 0.85 / 0.56 - 0.61 \\

coaster.\textit{N} & 0.68 - 0.85 - 0.93 / 0.85 - 0.94 & 
vamp.\textit{N} & 0.53 - 0.53 - 0.96 / 0.57 - 0.57 & 
wisely.\textit{RB} & 0.56 - 0.58 - 0.84 / 0.63 - 0.58 \\

teacher.\textit{N} & 0.4 - 0.91 - 0.96 / 0.91 - 0.96 
& tangible.\textit{JJ} & 0.52 - 0.57 - 0.74 / 0.57 - 0.57 & 
boxing.\textit{N} & 0.59 - 0.6 - 0.71 / 0.6 - 0.53 \\

blow.\textit{V} & 0.84 - 0.81 - 0.91 / 0.89 - 0.96 & 
jap.\textit{N} & 0.61 - 0.61 - 0.63 / 0.61 - 0.59 & podium.\textit{N} & 0.62 - 0.62 - 0.94 / 0.75 - 0.75 \\

fairy.\textit{N} & 0.69 - 1.0 - 1.0 / 1.0 - 1.0 & jelly.\textit{N} & 0.52 - 0.59 - 0.72 / 0.59 - 0.59 & 
ruler.\textit{N} & 0.48 - 0.48 - 0.91 / 0.42 - 0.7 \\

diversity.\textit{N} & 0.5 - 0.41 - 0.88 / 0.45 - 0.41 & 
proud.\textit{JJ} & 0.64 - 0.64 - 0.65 / 0.64 - 0.7 & goat.\textit{N} & 0.62 - 0.62 - 0.7 / 0.63 - 0.77 \\

postmortem.\textit{N} & 0.48 - 0.48 - 0.96 / 0.57 - 0.64 & extend.\textit{V} & 0.62 - 0.76 - 0.9 / 0.76 - 0.98 & bakery.\textit{N} & 0.45 - 0.45 - 0.73 / 0.38 - 0.5 \\

clamp.\textit{N} & 0.62 - 0.62 - 0.92 / 0.56 - 0.53 & 
remote.\textit{JJ} & 0.49 - 0.49 - 0.88 / 0.59 - 0.65 & 
dusty.\textit{JJ} & 0.79 - 0.85 - 0.97 / 0.82 - 0.88 \\

reconsider.\textit{V} & 0.61 - 0.61 - 0.6 / 0.61 - 0.58 & 
rethink.\textit{V} & 0.79 - 0.79 - 0.79 / 0.79 - 0.79 & 
agne.\textit{N} & 0.62 - 0.62 - 0.96 / 0.5 - 0.69 \\

knit.\textit{V} & 0.63 - 0.65 - 0.78 / 0.63 - 0.74 & 
conductor.\textit{N} & 0.47 - 0.5 - 0.91 / 0.61 - 0.82 & 
memorable.\textit{JJ} & 0.62 - 0.62 - 0.72 / 0.65 - 0.6 \\

medium.\textit{JJ} & 0.52 - 0.6 - 0.94 / 0.78 - 0.68 & 
elite.\textit{JJ} & 0.65 - 0.65 - 0.95 / 0.67 - 0.69 & 
holodeck.\textit{N} & 0.52 - 0.52 - 0.97 / 0.7 - 0.67 \\

bastard.\textit{N} & 0.38 - 0.68 - 0.95 / 0.76 - 0.89 & 
helm.\textit{N} & 0.81 - 0.83 - 0.79 / 0.81 - 0.78 & 
cradle.\textit{N} & 0.66 - 0.76 - 0.97 / 0.84 - 0.95 \\

vacant.\textit{JJ} & 0.69 - 0.77 - 0.99 / 0.82 - 0.95 & 
piss.\textit{V} & 0.69 - 0.69 - 0.69 / 0.69 - 0.7 & 
motto.\textit{N} & 0.48 - 0.46 - 0.46 / 0.48 - 0.45 \\

parliament.\textit{N} & 0.59 - 0.59 - 0.61 / 0.59 - 0.57 & 
hideout.\textit{N} & 0.56 - 0.56 - 0.56 / 0.56 - 0.46 & 
bond.\textit{N} & 0.63 - 0.66 - 0.9 / 0.61 - 0.98 \\

pretzel.\textit{N} & 0.7 - 0.7 - 0.69 / 0.7 - 0.62 & 
deek.\textit{N} & 0.61 - 0.61 - 0.62 / 0.61 - 0.48 & 
abnormal.\textit{JJ} & 0.66 - 0.61 - 0.67 / 0.66 - 0.59 \\

abrasion.\textit{N} & 0.61 - 0.39 - 0.6 / 0.61 - 0.71 & 
countess.\textit{N} & 0.58 - 0.58 - 0.58 / 0.58 - 0.48 & 
dioxide.\textit{N} & 0.66 - 0.66 - 0.68 / 0.66 - 0.72 \\

walkie.\textit{N} & 0.68 - 0.68 - 0.68 / 0.68 - 0.71 & 
tight.\textit{N} & 0.62 - 0.62 - 0.73 / 0.62 - 0.64 & 
invaluable.\textit{JJ} & 0.65 - 0.65 - 0.64 / 0.65 - 0.65 \\

tango.\textit{N} & 0.52 - 0.52 - 0.94 / 0.6 - 0.55 & 
pedal.\textit{N} & 0.53 - 0.49 - 0.92 / 0.53 - 0.63 & 
petal.\textit{N} & 0.56 - 0.83 - 0.92 / 0.78 - 0.81 \\

expand.\textit{V} & 0.54 - 0.76 - 0.97 / 0.84 - 0.95 & 
envy.\textit{N} & 0.56 - 0.56 - 0.92 / 0.64 - 0.71 & 
partially.\textit{RB} & 0.57 - 0.66 - 0.87 / 0.55 - 0.66 \\

resourceful.\textit{JJ} & 0.6 - 0.6 - 0.6 / 0.6 - 0.51 & 
heinous.\textit{JJ} & 0.72 - 0.28 - 0.84 / 0.88 - 0.84 & 
contusion.\textit{N} & 0.55 - 0.53 - 0.9 / 0.61 - 0.58 \\

seeker.\textit{N} & 0.55 - 0.55 - 0.58 / 0.55 - 0.59  
weep.\textit{V} & 0.62 - 0.67 - 0.61 / 0.62 - 0.95 & 
jewel.\textit{N} & 0.7 - 0.3 - 0.7 / 0.7 - 0.7 \\

algae.\textit{N} & 0.55 - 0.67 - 0.85 / 0.58 - 0.58 & 
newlywed.\textit{N} & 0.62 - 0.62 - 0.6 / 0.62 - 0.79 & 
barge.\textit{N} & 0.56 - 0.56 - 0.91 / 0.59 - 0.75 \\

fatigue.\textit{N} & 0.52 - 0.45 - 0.88 / 0.5 - 0.59 & 
farmhouse.\textit{N} & 0.54 - 0.54 - 0.64 / 0.54 - 0.46 & 
imply.\textit{V} & 0.72 - 0.72 - 0.68 / 0.72 - 0.9\\

gator.\textit{N} & 0.73 - 0.76 - 0.73 / 0.73 - 0.84 & 
outskirt.\textit{N} & 0.52 - 0.52 - 0.82 / 0.52 - 0.76 & 
gel.\textit{N} & 0.5 - 0.73 - 0.81 / 0.75 - 0.68 \\

riddance.\textit{N} & 0.52 - 0.52 - 0.7 / 0.38 - 0.76 & 
infectious.\textit{JJ} & 0.54 - 0.57 - 0.79 / 0.57 - 0.54 & 
chord.\textit{N} & 0.62 - 0.72 - 0.8 / 0.88 - 0.66 \\

hide.\textit{V} & 0.75 - 0.75 - 0.98 / 0.86 - 0.82 & 
conquer.\textit{V} & 0.56 - 1.0 - 1.0 / 1.0 - 1.0 & 
compression.\textit{N} & 0.76 - 0.91 - 0.91 / 0.88 - 0.85 \\

cunning.\textit{JJ} & 0.56 - 0.56 - 0.87 / 0.62 - 0.66 & 
plague.\textit{V} & 0.62 - 0.62 - 0.98 / 0.94 - 1.0 & 
morbid.\textit{N} & 0.6 - 0.52 - 0.63 / 0.6 - 0.56 \\

particle.\textit{N} & 0.55 - 0.92 - 0.97 / 0.92 - 1.0 & 
theo.\textit{N} & 0.64 - 0.64 - 0.96 / 0.62 - 0.5 & 
femur.\textit{N} & 0.61 - 0.61 - 0.61 / 0.61 - 0.61 \\

commit.\textit{V} & 0.92 - 0.93 - 0.93 / 0.92 - 0.99 & 
rivalry.\textit{N} & 0.58 - 0.5 - 0.84 / 0.54 - 0.54 & 
pyjama.\textit{N} & 0.62 - 0.62 - 0.62 / 0.62 - 0.66 \\
scatter.\textit{V} & 0.59 - 0.63 - 0.61 / 0.59 - 0.67 & 
donor.\textit{N} & 0.53 - 0.88 - 0.94 / 0.91 - 0.94 & 
smither.\textit{N} & 0.6 - 0.6 - 0.97 / 0.47 - 0.43 \\ 
yuan.\textit{N} & 0.59 - 0.59 - 0.88 / 0.73 - 0.73 & 
autopsy.\textit{N} & 0.56 - 0.56 - 0.92 / 0.6 - 0.52 & 
gullible.\textit{JJ} & 0.52 - 0.41 - 0.89 / 0.37 - 0.63 \\ 
absolution.\textit{N} & 0.59 - 0.59 - 0.9 / 0.48 - 0.69 & 
modesty.\textit{N} & 0.58 - 0.75 - 0.83 / 0.71 - 0.71 & 
handgun.\textit{N} & 0.58 - 0.58 - 0.59 / 0.58 - 0.42 \\
firecracker.\textit{N} & 0.59 - 0.64 - 0.86 / 0.5 - 0.73 & 
honorary.\textit{JJ} & 0.58 - 0.67 - 0.91 / 0.67 - 0.92 & 
rod.\textit{N} & 0.66 - 0.66 - 0.98 / 0.74 - 0.83 \\
kneecap.\textit{N} & 0.54 - 0.54 - 0.95 / 0.62 - 0.69 & 
insubordination.\textit{N} & 0.56 - 0.56 - 0.89 / 0.32 - 0.64 & 
poacher.\textit{N} & 0.52 - 0.52 - 0.98 / 0.43 - 0.43 \\
railing.\textit{N} & 0.56 - 0.56 - 0.61 / 0.56 - 0.72 & 
tremor.\textit{N} & 0.59 - 0.55 - 0.82 / 0.64 - 0.86 & 
coalition.\textit{N} & 0.52 - 0.43 - 0.83 / 0.61 - 0.52 \\ 
carnage.\textit{N} & 0.55 - 0.52 - 0.56 / 0.55 - 0.59 & 
memento.\textit{N} & 0.59 - 0.38 - 0.84 / 0.38 - 0.59 \\

\midrule
 \multicolumn{3}{r|}{\textbf{Overall Average:}} & 58.56 - 63.79 - 66.46 - 71.74 \\
 \bottomrule
\end{tabular}
    }
    \caption{Part-3: Lexical selection model test accuracies for a English-Greek words.}
   \label{tab:results-3}
     \end{center}
\end{table*}

\end{document}